\documentclass[conference]{IEEEtran}
\IEEEoverridecommandlockouts
\usepackage{cite}
\usepackage{amsmath,amssymb,amsfonts}
\usepackage{algorithmic}
\usepackage{graphicx}
\usepackage{textcomp}
\usepackage{subfigure}
\usepackage{algorithm}
\usepackage{algorithmic}
\usepackage{xcolor}
\def\BibTeX{{\rm B\kern-.05em{\sc i\kern-.025em b}\kern-.08em
    T\kern-.1667em\lower.7ex\hbox{E}\kern-.125emX}}
\begin{document}

\title{An End-to-End Joint Unsupervised Learning of Deep Model and Pseudo-Classes for Remote Sensing Scene Representation
\thanks{This work was
supported in part by the NSF of China under Grant
61671456 and Grant 61271439, FANEDD under Grant 201243, and
 Program for New Century Excellent Talents in University under
Grant NECT-13-0164.}
}

\author{\IEEEauthorblockN{Zhiqiang Gong, Ping Zhong, Weidong Hu, Fang Liu, and Bingwei Hui}
\IEEEauthorblockA{\textit{National Key Laboratory of Science and Technology on ATR} \\
\textit{National University of Defense Technology}\\
Changsha 410073, China \\
E-mail: zhongping@nudt.edu.cn}

}

\maketitle

\begin{abstract}
This work develops a novel end-to-end deep unsupervised learning method based on convolutional neural network (CNN) with pseudo-classes for remote sensing scene representation.
First, we introduce center points as the centers of the pseudo classes and the training samples can be allocated with pseudo labels based on the center points.
Therefore, the CNN model, which is used to extract features from the scenes, can be trained supervised with the pseudo labels.    Moreover, a pseudo-center loss is developed to decrease the variance between the samples and the corresponding pseudo center point. The pseudo-center loss is important since it can update both the center points with the training samples and the CNN model with the center points in the training process simultaneously.  Finally, joint learning of the pseudo-center loss and the pseudo softmax loss which is formulated with the samples and the pseudo labels is developed for unsupervised remote sensing scene representation to obtain discriminative representations from the scenes.
Experiments are conducted over two commonly used remote sensing scene datasets to validate the effectiveness of the proposed method and the experimental results show the superiority of the proposed method when compared with other state-of-the-art methods.
\end{abstract}

\begin{IEEEkeywords}
Unsupervised Learning, Pseudo-Class, End-to-End Learning, Convolutional Neural Network (CNN), Remote Sensing Scene Representation
\end{IEEEkeywords}

\section{Introduction}

Nowadays, high resolution images from the new and the advanced space-borne or aerial-borne sensors contain abundant spatial and spectral information, which could provide helpful information for many military or civilian applications. However, efficient representation and recognition of the remote sensing scenes tend to be a challenging problem since labelling is generally time-consuming and sometimes infeasible \cite{b1}. Therefore, unsupervised learning methods tend to be a hot topic to extract discriminative features without the labels of the scenes. The general unsupervised learning methods, such as SIFT \cite{sift} and LBP \cite{lbp}, captures the geometrical information, salient points or the textural information from the scenes. However, the complex arrangements in the scenes, the large inner-class variance and low inter-class variance between different scenes make it difficult to discriminate the scenes from overlapping classes with these low-level features.

In recent years, deep learning methods have shown powerful ability to extract high-level features from the objects. Many deep learning-based unsupervised representations have been developed. It can be divided into four categories: the self-supervised learning approaches which tries to implement a supervised learning with pseudo labels which is created in an unsupervised way, the reconstruction-based methods, the generative adversarial network(GAN)-based methods, and the natural rule motivated loss-function methods \cite{b1}. In the first one, the way to construct the pseudo classes plays an important role since they directly affect the training efficiency for the remote sensing scenes. This work will focus on developing an efficient end-to-end unsupervised learning from the self-supervised learning way.

Prior works mainly construct the pseudo classes from three aspects. The first one is to extract the image patches from the scenes and further construct the pseudo classes together with the transformations of the image patches \cite{b8}. Some other works take advantage of the selective search method to extract patches with multi-scale information from the scene to form the pseudo classes \cite{b9}. Another work makes use of the inner-correlation between the image patches in a scene image to form the pseudo classes \cite{b10}. These pseudo classes are generated according to the special requirements of different tasks and the pseudo classes are usually fixed in the training process. However, these prior works mainly take advantage of image patches extracted from the scenes to form the pseudo classes which ignores some important information in the scene image. This would limit the performance of the learned model to extract discriminative features from the scenes.

\begin{figure*}[htbp]
\centerline{
{\includegraphics[width=0.9\textwidth]{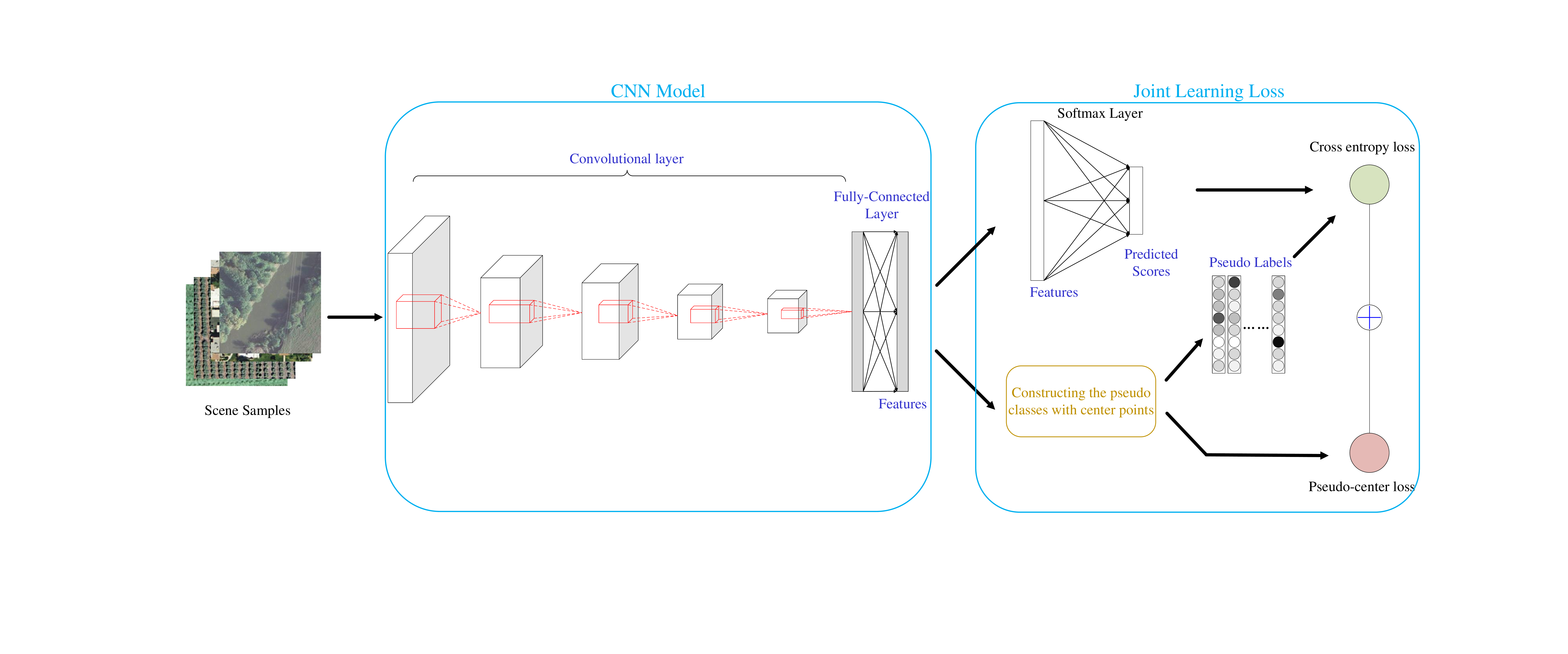}}}
\caption{Flowchart of the proposed method for unsupervised learning of remote sensing scenes. The CNN model is used to extract deep features from the remote sensing scenes. The joint learning loss tries to update both the center points and the CNN model simultaneously. Therefore, the pseudo classes are updated with the update of the center points  and the center points tend to describe the centers of different classes. Then, the learned model can be better fit for the remote sensing scenes and can discriminate scenes from different classes.}
\label{fig:flowchart}
\end{figure*}

To overcome this problem, this work denotes the center points to represent the pseudo classes. Based on the center points, we allocate the training samples to the nearest pseudo class. Motivated by the prior work \cite{center}, a novel pseudo center loss is formulated with the training samples and the corresponding center point the samples belong to. The center points would be updated with the pseudo center loss in the training process and the pseudo label of each training sample would also be changed with the updated center points in the unsupervised training process.

To take advantage of both the deep representation and the pseudo center loss, this work develops a novel end-to-end  joint unsupervised learning of deep model and pseudo classes for the remote sensing scenes, which jointly learns the pseudo center loss and the pseudo softmax loss. The pseudo softmax loss which is formulated with the pseudo labels is used to update the deep model.  Through updating the deep CNN model and the center points which represent the pseudo classes simultaneously, the pseudo classes would be close to the real classes and the learned features from the deep model would be more discriminative.

The rest of the paper is arranged as follows. In section II, we briefly introduce the general convolutional neural network,  develop the end-to-end deep unsupervised learning method with pseudo-classes for remote sensing scene representation, and introduces the implementation of the proposed method in detail. Details of our experiments and results are presented in Section III. Section IV concludes the paper with some discussions.

\section{Proposed Method}

In this section, we first briefly introduce the general convolutional neural network (CNN), and then develops the pseudo center loss with the pseudo-classes for unsupervised learning of remote sensing scenes, and then the joint learning method for unsupervised learning is developed, and finally we present the implementation of the proposed method for remote sensing scene representation.

Let us denote ${\bf x}_i (i=1, 2, \cdots, N)$ as the samples from the remote sensing scenes and $N$ is the number of the unlabelled scenes.

\subsection{General Convolutional Neural Network (CNN)}

Deep learning-based method, such as Convolutional Neural Network (CNN), Deep Belief Network (DBN), have shown their impressive performance for remote sensing scene representation \cite{brazilian}. Among these methods, CNNs which can extract both the local and global features from the scenes have been widely used in the literature of remote sensing \cite{gong,brazilian}.   As Fig. \ref{fig:flowchart} shows,  this work chooses the CNN model to extract features from the scenes.

The general CNNs consist of layers of many types, such as the convolutional layer, fully connected layer, pooling layer, ReLU layer, loss layer.
It can be looked as the parallel of these layers where the output of the former layer is performed as the input of the current layer. Denote ${\bf s}^k$ as the features learned from the $k^{th}$ layer, and then the features ${\bf s}^{k+1}$ that obtained from the $k+1^{th}$ layer can be calculated by
\begin{equation}\label{eq:1_1}
  {\bf s}^{k+1}=f(W_k{\bf s}^k+{\bf b}_k)
\end{equation}
where $W_k$ and ${\bf b}_k$ represent the parameters and the bias in the $k^{th}$ layer. $f(\cdot)$ denotes the nonlinear activation function.

To accurately train the deep model,
the training batch, which denotes a set of samples that train the deep model simultaneously, is usually used in the training process. In addition,
the softmax loss, which consists of softmax layer and cross entropy loss, is generally used for the training of the CNN model.

\subsection{Pseudo Center Loss with Pseudo-Classes} \label{subsec:center_loss}

Denote ${\bf c}_i (i=1,2,\cdots, \Lambda)$ as the center point where each center point represents a pseudo class and $\Lambda$ represents the number of the pseudo classes.
To provide the pseudo labels to the unlabelled samples, the key process is to formulate the variance between the samples and different pseudo classes.

\begin{figure}[htbp]
\centerline{
{\includegraphics[width=0.45\textwidth]{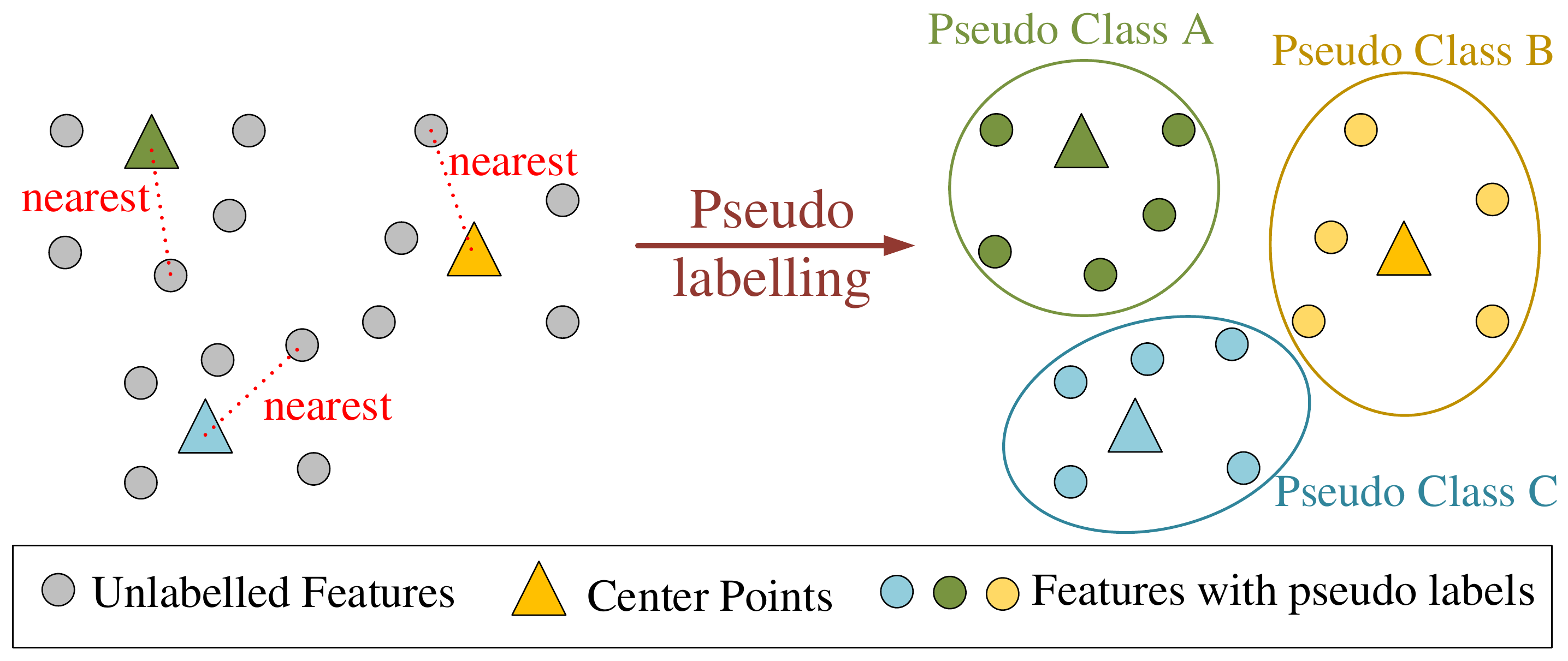}}}
\caption{The process of formulating the pseudo classes with the training samples in the training batch based on the center points. The sample is allocated with the pseudo label from the nearest center point. It should be noted that the pseudo classes would be dynamic changed with the update of the center points and the CNN model.}
\label{fig:labelling}
\end{figure}

Given a training batch $B$. For each sample ${\bf x}_i\in B$,
denote $\varphi({\bf x}_i)$ as the features extracted from the CNN model.
Since the center points ${\bf c}_i (i=1,2,\cdots,\Lambda)$ are constructed to represent different classes, the pseudo class each sample in the training batch $B$ belongs to can be calculated by
\begin{equation}\label{eq:2}
  z_i=\arg\min_l \|{\bf c}_l-\varphi({\bf x}_i)\|, (l\in \{1,2,\cdots, \Lambda\}),
\end{equation}
The process for allocating pseudo labels to the training samples in the batch is shown in Fig. \ref{fig:labelling}. Since the pseudo classes and the CNN model are dynamic changed, the pseudo label of each sample is changed in the training process.

Since the construction of the pseudo classes is related to the center points, the update of the center points can significantly affect the effectiveness of the training process as well as the performance of the representation for the scenes. Motivated by \cite{center}, this work formulates the pseudo center loss with the samples in the batch to update the center points. The pseudo center loss tries to encourage the center points to approach the training samples in the pseudo class and it can be formulated as
\begin{equation}\label{eq:4}
  L_c=\sum_{i=1}^{|B|} \|{\bf c}_{z_i}-\varphi({\bf x}_i)\|^2
\end{equation}
where $z_i$ is the pseudo label of ${\bf x}_i$ calculated from Eq. \ref{eq:2}.
In the training process, the $L_c$ is used to update both the parameters in the CNN model and the center points.

\subsection{Joint Learning Loss for Unsupervised Learning of Remote Sensing Scene Representation} \label{subsec:proposed}

As general self-supervised learning approaches, this work uses the softmax loss to supervised learn the CNN model with the pseudo classes.  The pseudo softmax loss formulated by the samples in different pseudo classes  can be calculated by
\begin{equation}\label{eq:3}
  L_s=-\sum_{i=1}^{|B|}\log\frac{e^{W^T_{0,{z}_i}\varphi({{\bf x}_i})+b_{0,z_i}}}{\sum_{j=1}^{\Lambda}e^{W^T_{0,j}\varphi({{\bf x}_i})+b_{0,j}}}
\end{equation}
where $W_0=[W_{0,1},W_{0,2},\cdots,W_{0,\Lambda}]$, ${\bf b}_0=[b_{0,1},b_{0,2},\cdots,b_{0,\Lambda}]$ represent the parameters and the bias term in Softmax layer, respectively.
The $L_s$ is used to calculate the penalization between the predicted scores over the pseudo classes with the pseudo labels by Eq. \ref{eq:2}.



Considering the merits of the CNN model and the center points which are used to formulate the pseudo classes, this work develops a novel joint learning loss of the pseudo classes and the deep model. It can be formulated as
\begin{equation}\label{eq:5}
\begin{aligned}
  L=&L_s+\lambda L_c \\
  =&-\sum_{i=1}^{|B|}\log\frac{e^{W^T_{0,{z}_i}\varphi({\bf x}_i)+b_{0,z_i}}}{\sum_{j=1}^{\Lambda}e^{W^T_{0,j}\varphi({\bf x}_i)+b_{0,j}}}+\lambda\sum_{i=1}^{|B|} \|{\bf c}_{z_i}-\varphi({\bf x}_i)\|^2
\end{aligned}
\end{equation}
where $\lambda$ is a positive value which denotes the tradeoff between the pseudo softmax loss and the pseudo center loss. The developed pseudo classes-based loss can jointly learns the deep model and the pseudo classes simultaneously. Therefore, the learned model can be more fit for the remote sensing scenes and could discriminate scenes with great similarity from different classes.

\subsection{Implementation of the Proposed Method}

The proposed unsupervised learning process can be trained end-to-end by the stochastic gradient descent (SGD).  According to the characteristics of the back propagation of the deep model \cite{back_propagation}, the main problem is to calculate the partial of the joint learning loss w.r.t. ${\bf x}_i$. More importantly, in this work, the update of the pseudo classes should also be implemented by calculating the partial of the joint learning loss w.r.t. the center points.

The partial of the pseudo softmax loss $L_s$ w.r.t. ${\bf x}_i$ can be calculated as Caffe which is the deep learning framework used in the  experiments \cite{caffe}. The partial of the pseudo center loss can be calculated as \cite{center} shows. Therefore, the partial of the proposed joint learning loss w.r.t. ${\bf x}_i$ can be calculated by
\begin{equation}\label{eq:6}
\frac{\partial L}{\partial \varphi({\bf x}_i)}=\frac{\partial L_s}{\partial \varphi({\bf x}_i)}+2\lambda (\varphi({\bf x}_i) -{\bf c}_{z_i}),
\end{equation}
where $z_i$ is the pseudo label of ${\bf x}_i$. This is used for the update of the parameters in the CNN model.

\begin{algorithm}[htbp]
\caption{Implementation of the unsupervised learning method}
\begin{algorithmic}[1]
\REQUIRE ${{\bf x}_i}(i=1,2,\cdots,N)$, $\theta_k=\{W_k, {\bf b}_k\}$ as the parameter of the $k^{th}$ convolutional layer, $W_0$ as the parameters and ${\bf b}_0$ is the bias term  in Softmax layer, hyperparameter $\lambda$, learning rate $lr$, the number of pseudo classes $\Lambda$.
\ENSURE  $\theta_k$
\STATE Initialize $\theta_k$ in $k^{th}$ convolution layer where $W_k$ is initialized from Gaussian distribution with standard deviation of 0.01 and $b_k$ is set to 0.  Initialize the center point ${\bf c}_i(i=1,2,\cdots,\Lambda)$ where ${\bf c}_i$ is filled with 0.
\WHILE{not converge}
\STATE $t \leftarrow t+1$.
\STATE Construct the training batch $B^t$.
\STATE Obtain the features $\varphi({\bf x}^t_i)$ of ${\bf x}^t_i\in B^t$ from CNN model with $\theta^t_{k}$.
\STATE Obtain the pseudo label $z^t_i$ of ${\bf x}^t_i\in B^t$ as Eq. \ref{eq:2} shows.
\STATE Compute the pseudo center loss with the pseudo labels of samples by $L^t_c=\sum_{i=1}^{|B^t|} \|{\bf c}^t_{z^t_i}-\varphi({\bf x}^t_i)\|^2$.
\STATE Compute the joint learning loss by $L^t=L^t_s+\lambda L^t_c$ where $L^t_s$ is calculated as Eq. \ref{eq:3}.
\STATE Compute the deviation ${{L^t}}$ w.r.t. $\varphi({{\bf x}^t_i})$ in $B^t$ by $\displaystyle{\frac{\partial {L^t}}{\partial \varphi({\bf x}^t_i)}=\frac{\partial {L_s^t}}{\partial \varphi({\bf x}^t_i)}+2\lambda (\varphi({\bf x}^t_i) -{\bf c}^t_{z^t_i})}$.
\STATE Compute the deviation $L^t$ w.r.t ${\bf c}^t_j$ by $\displaystyle{\frac{\partial L^t}{\partial {\bf c}^t_j}=}$ $\displaystyle{2\lambda \sum_{{\bf x}^t_i\in B^t}I(z^t_i=j)({\bf c}^t_j-\varphi({\bf x}^t_{i}))}$.
\STATE Update the parameters $W$ by $\displaystyle{W^{t+1}=}$ $\displaystyle{W^t-lr\times \frac{\partial L^t}{\partial W^t}= W^t-lr\times \frac{\partial L_s^t}{\partial W^t}}$.
\STATE Update the parameters $\theta_k$ of $k^{th}$ layer by $\displaystyle{\theta^{t+1}_k=}$ $\displaystyle{\theta^{t}_k-lr\times \frac{\partial L^t}{\partial \theta^t_k}=\theta^{t}_k-lr\times \sum_{i=1}^{|B|}\frac{\partial L^t}{\partial \varphi({\bf x}^t_i)}\times \frac{\partial \varphi({\bf x}^t_i)}{\partial \theta^t_k}}$.
\STATE Update the center points ${\bf c}_j$ by $\displaystyle{{\bf c}^{t+1}_j={\bf c}^{t}_j-lr\times  \frac{\partial L^t_c}{\partial {\bf c}^t_j}}$.
\ENDWHILE
\RETURN $\theta_k$
\end{algorithmic}
\label{algorithm:1}
\end{algorithm}

In addition, the partial of the proposed joint learning loss w.r.t. ${\bf c}_j$ can be calculated as
\begin{equation}\label{eq:7}
  \frac{\partial L}{\partial {\bf c}_j}=2\lambda \sum_{{\bf x}_i\in B}I(z_i=j)({\bf c}_j-\varphi({\bf x}_{i})).
\end{equation}
where $I(\cdot)$ represents the indicative function. $\displaystyle{\frac{\partial L}{\partial {\bf c}_j}}$, which is used to update the center points in the training process, can adjust the pseudo classes to the real one and make the learned features from the scenes be discriminative.
The overall unsupervised learning framework is given in Algorithm \ref{algorithm:1}.



\section{Experimental Results}

\subsection{Experimental Setup}
\begin{figure}[htbp]

\centerline{
\subfigure[]{\includegraphics[width=0.063\textwidth]{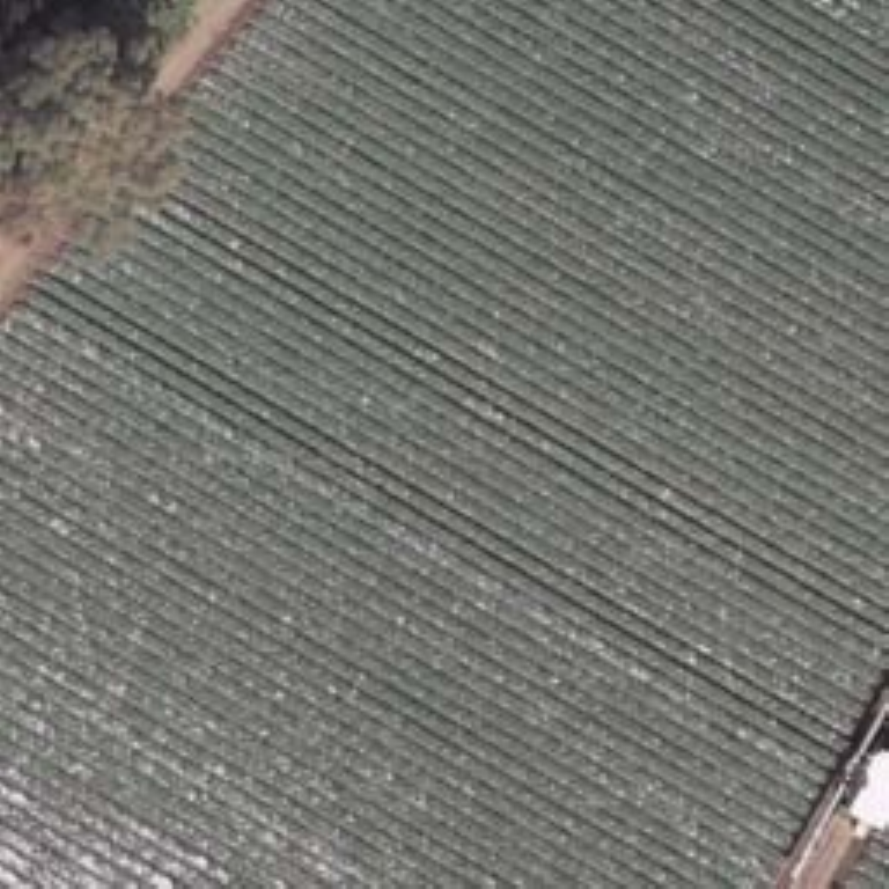}}
\subfigure[]{\includegraphics[width=0.063\textwidth]{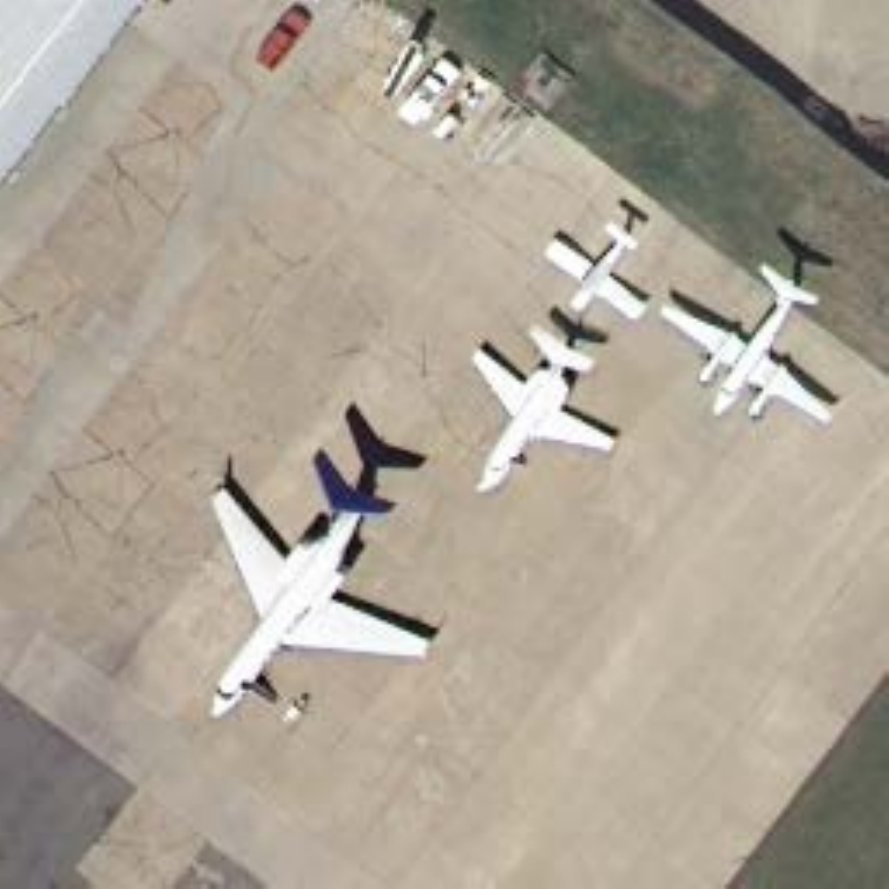}}
\subfigure[]{\includegraphics[width=0.063\textwidth]{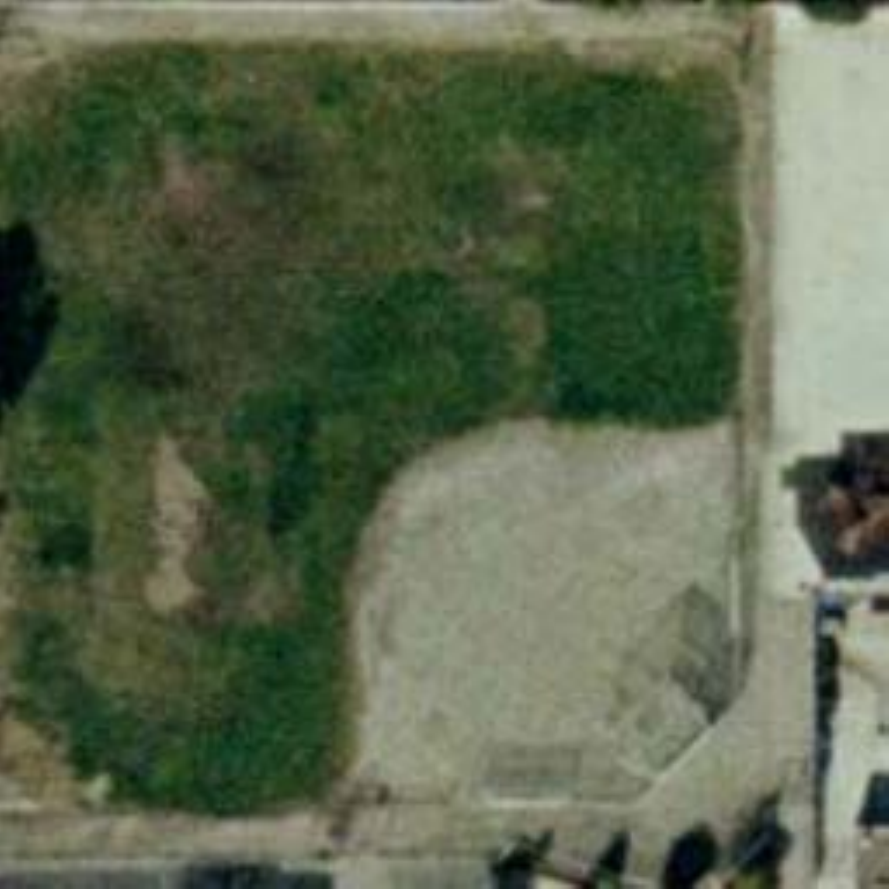}}
\subfigure[]{\includegraphics[width=0.063\textwidth]{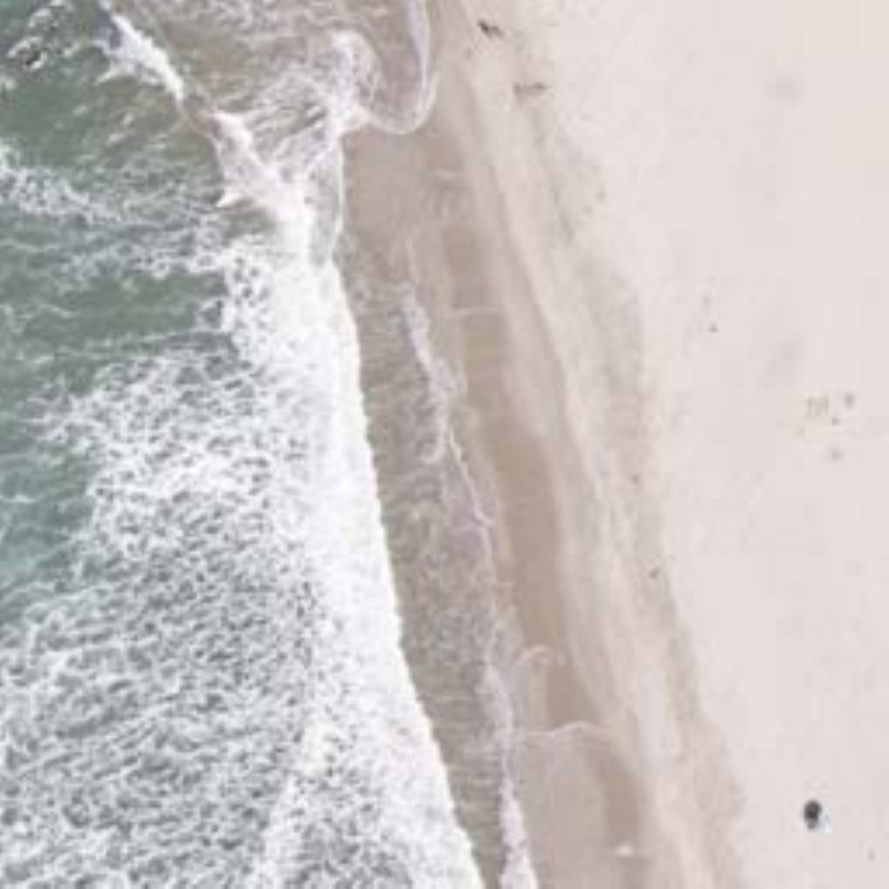}}
\subfigure[]{\includegraphics[width=0.063\textwidth]{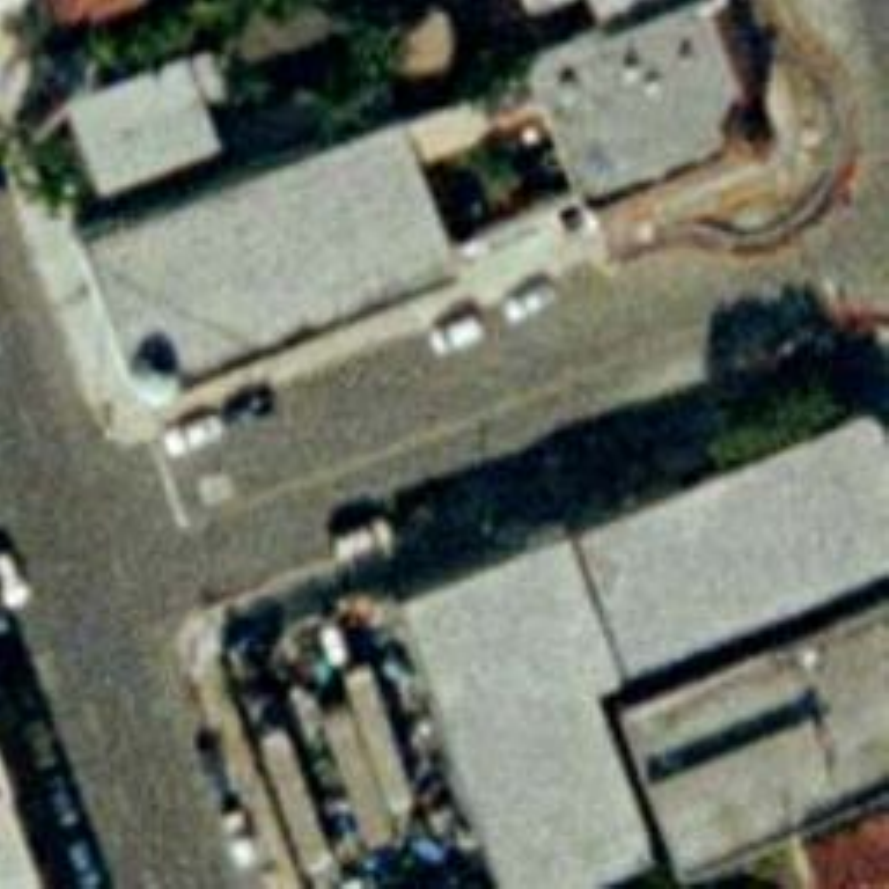}}
\subfigure[]{\includegraphics[width=0.063\textwidth]{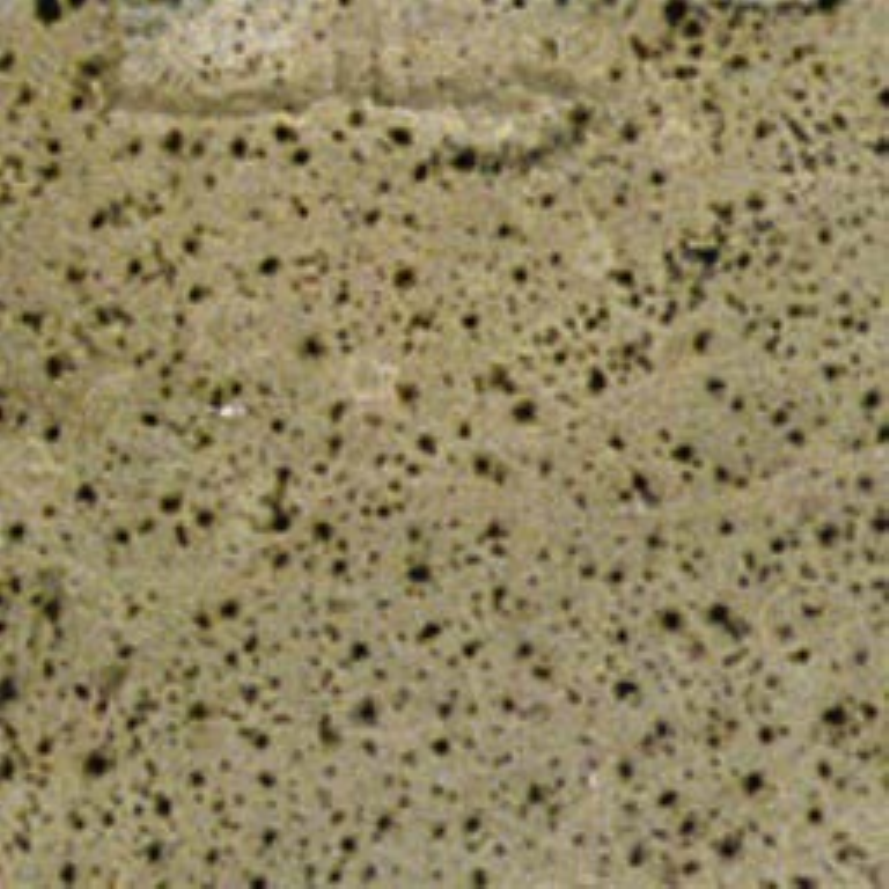}}
\subfigure[]{\includegraphics[width=0.063\textwidth]{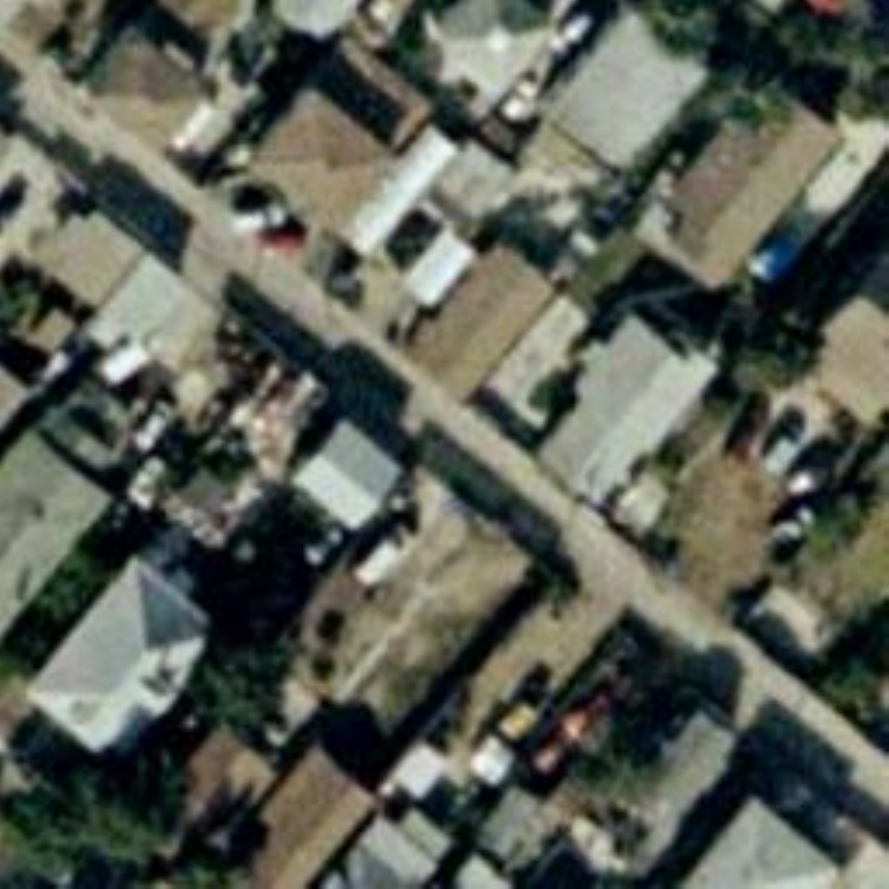}}}
\centerline{
\subfigure[]{\includegraphics[width=0.063\textwidth]{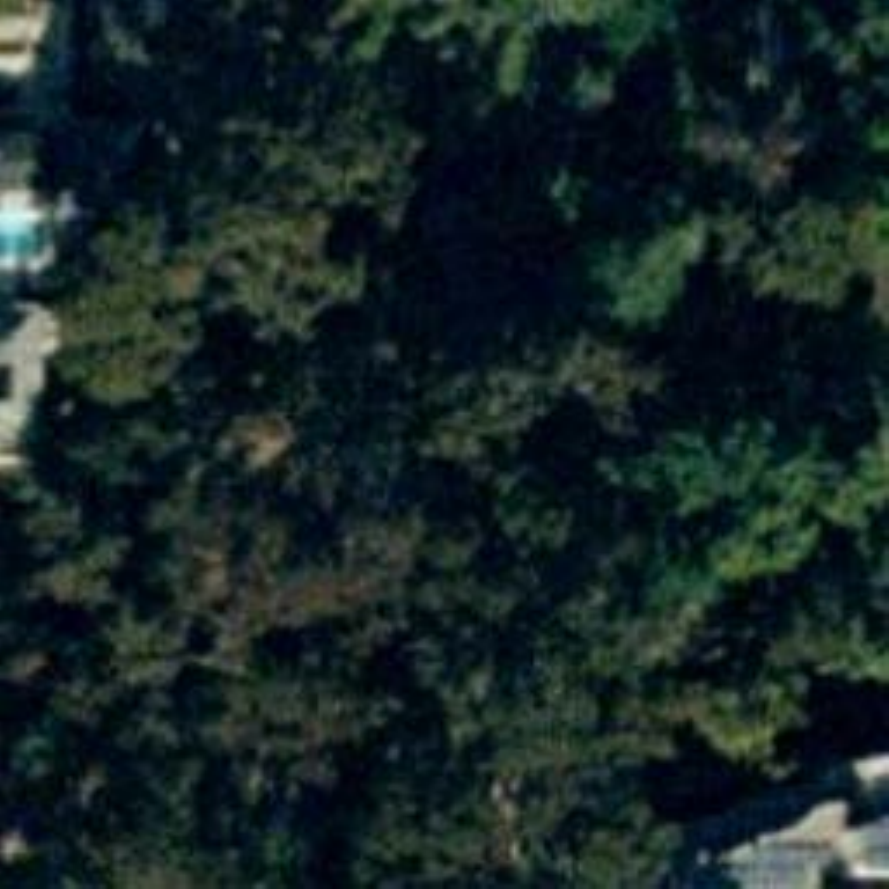}}
\subfigure[]{\includegraphics[width=0.063\textwidth]{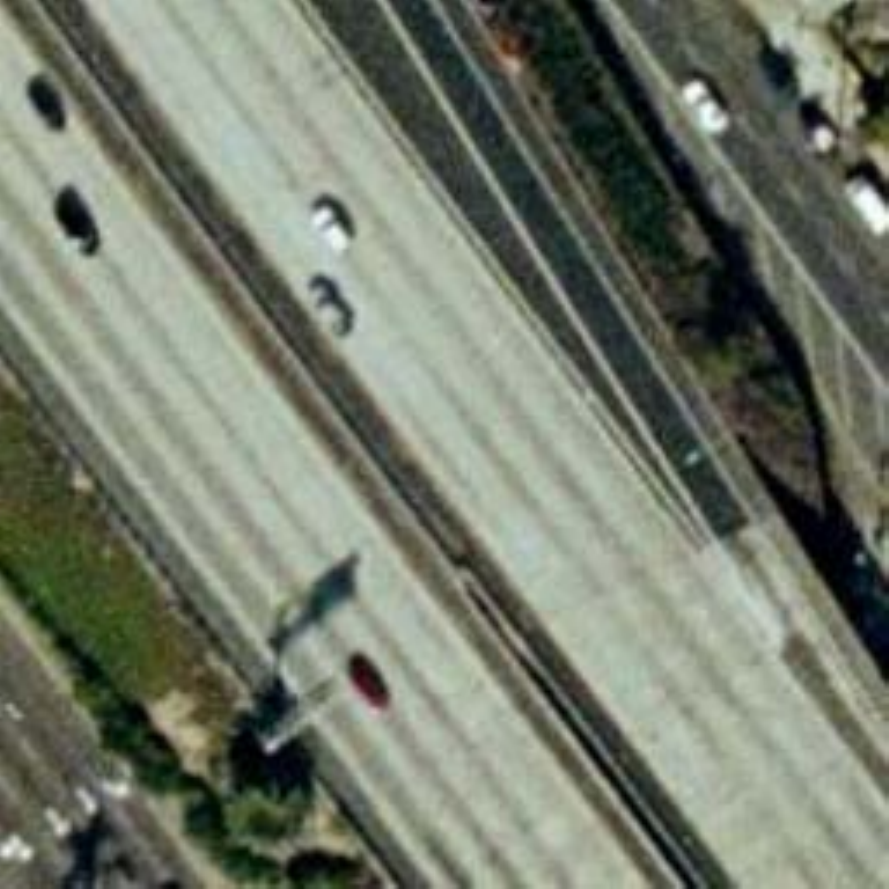}}
\subfigure[]{\includegraphics[width=0.063\textwidth]{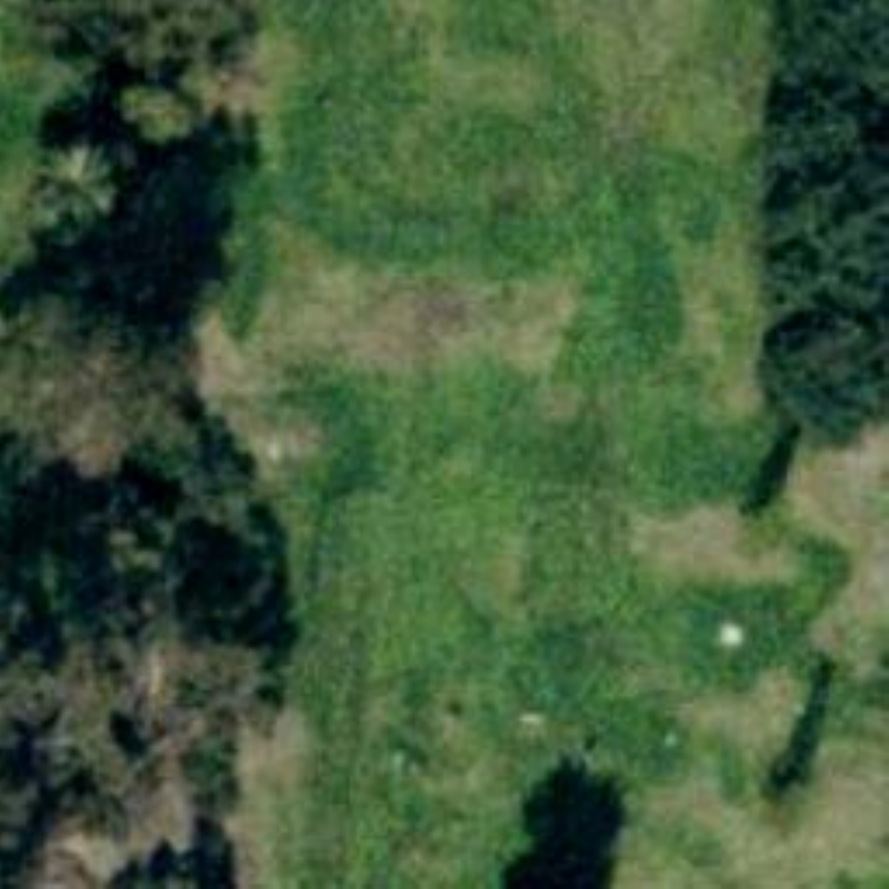}}
\subfigure[]{\includegraphics[width=0.063\textwidth]{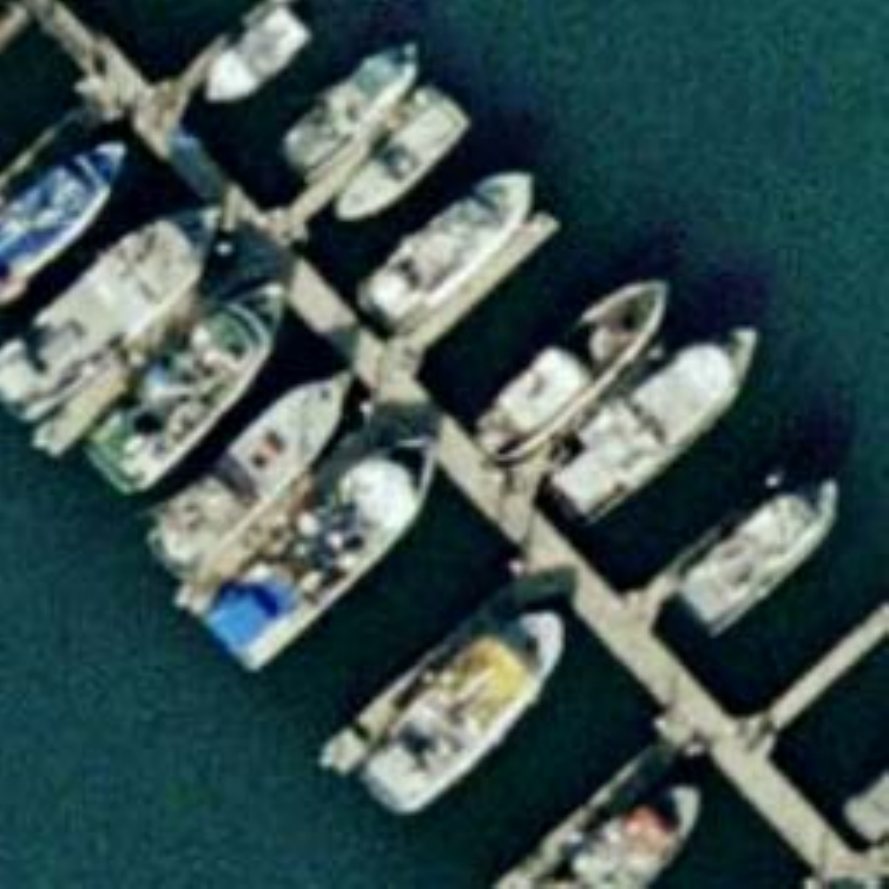}}
\subfigure[]{\includegraphics[width=0.063\textwidth]{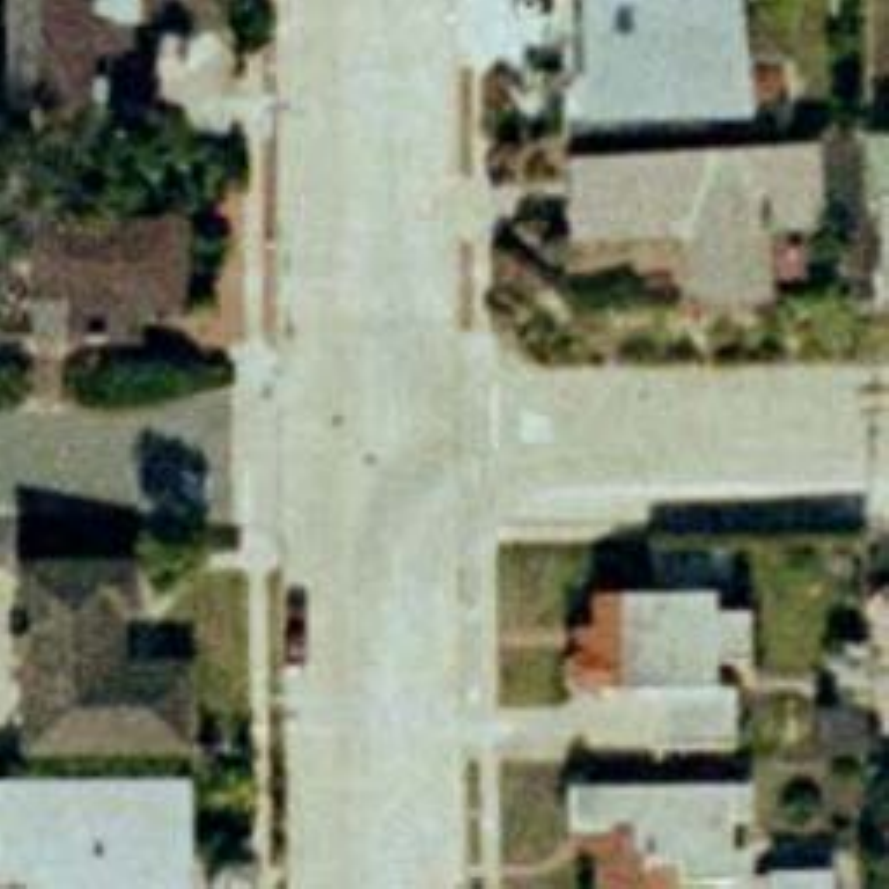}}
\subfigure[]{\includegraphics[width=0.063\textwidth]{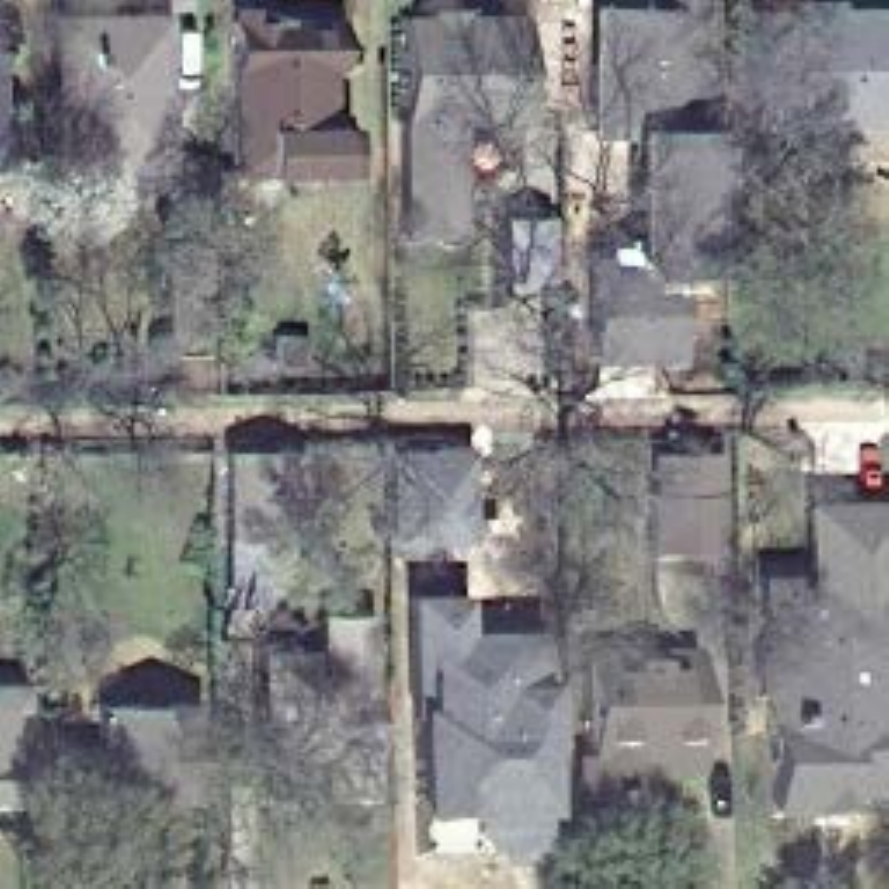}}
\subfigure[]{\includegraphics[width=0.063\textwidth]{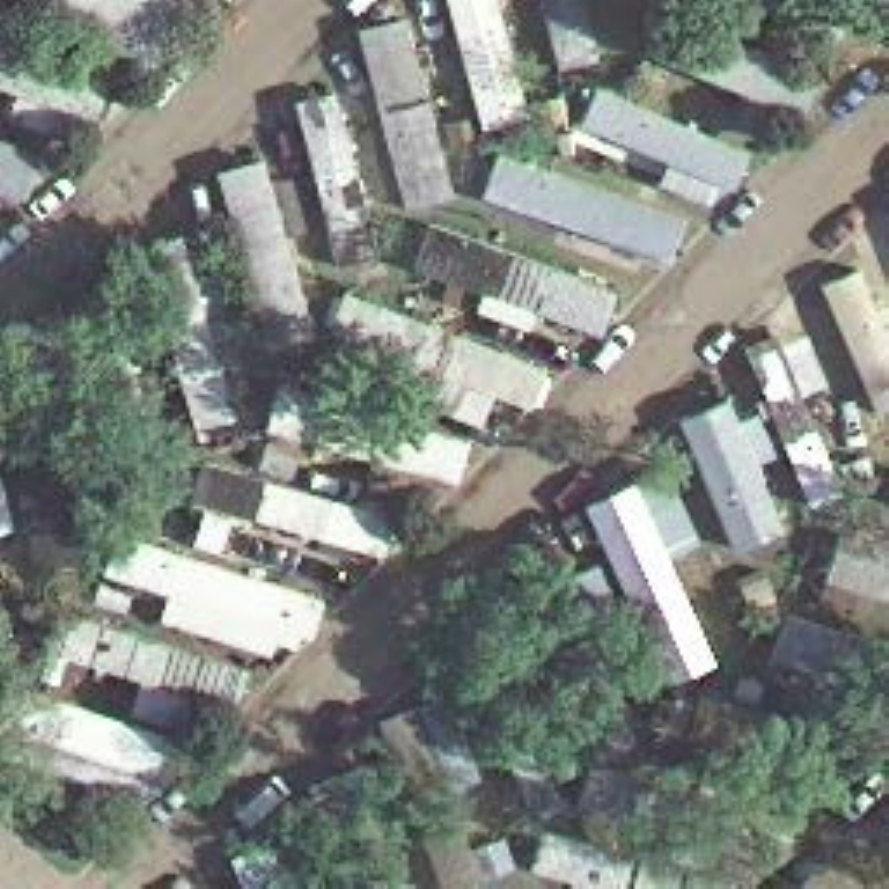}}}
\centerline{
\subfigure[]{\includegraphics[width=0.063\textwidth]{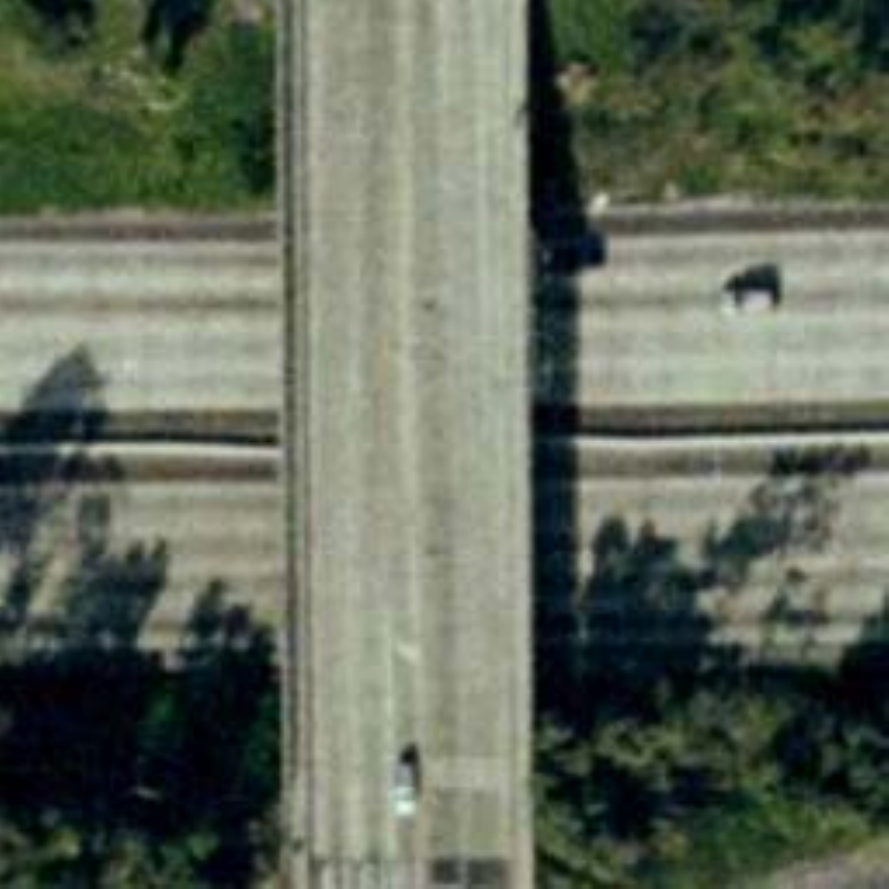}}
\subfigure[]{\includegraphics[width=0.063\textwidth]{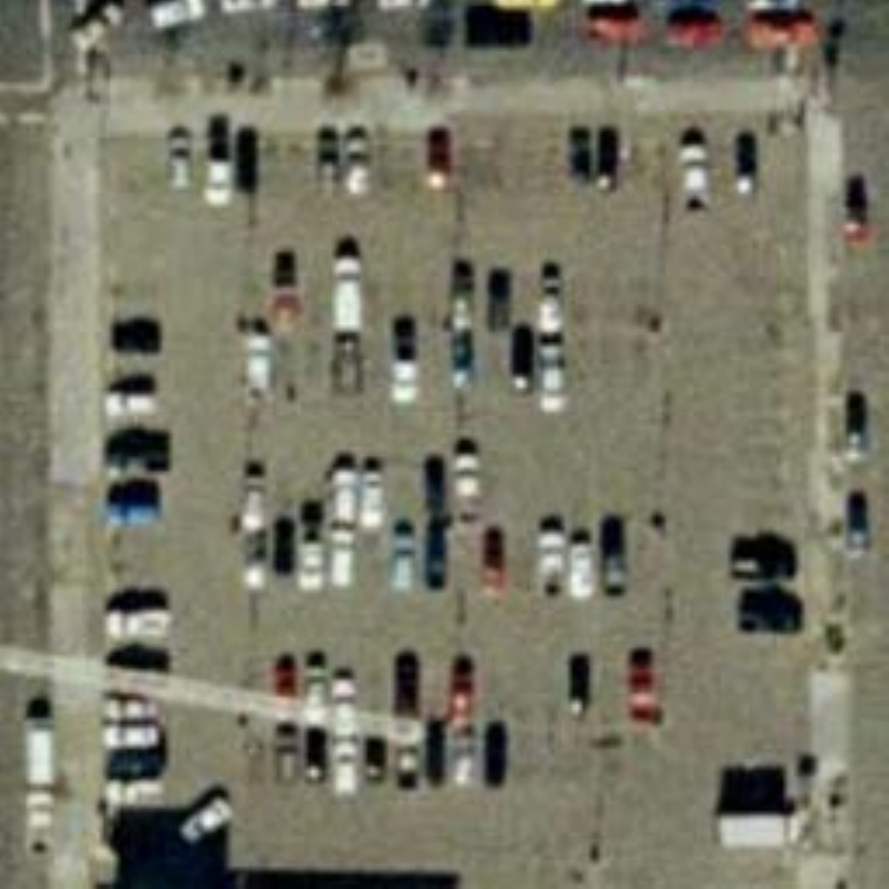}}
\subfigure[]{\includegraphics[width=0.063\textwidth]{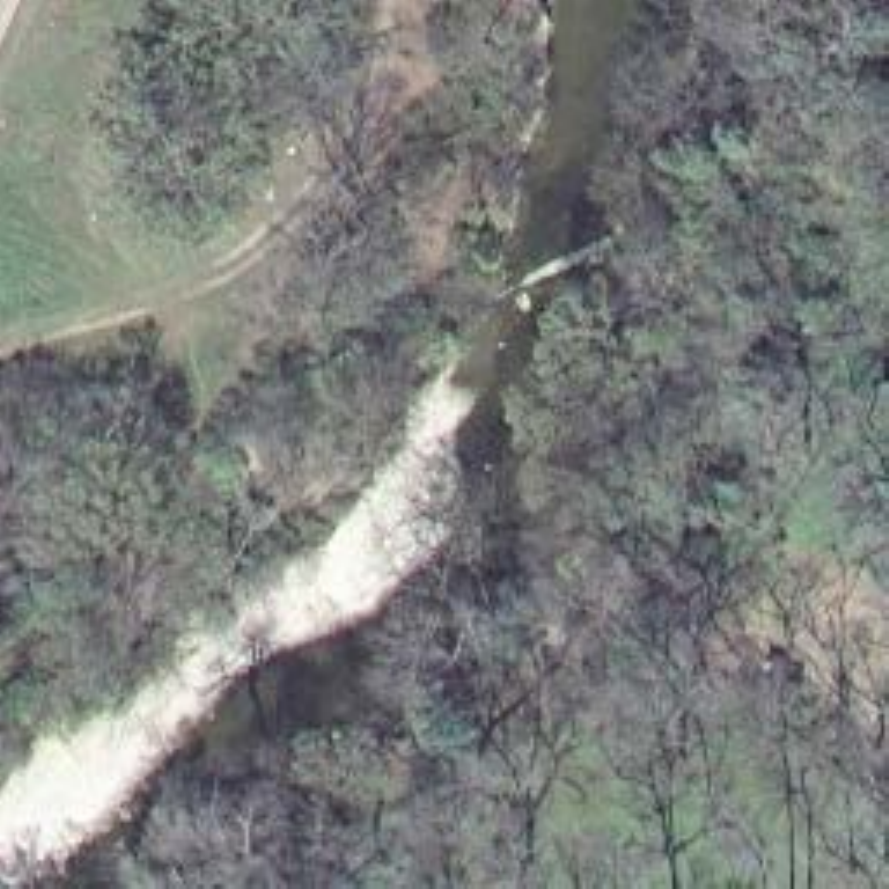}}
\subfigure[]{\includegraphics[width=0.063\textwidth]{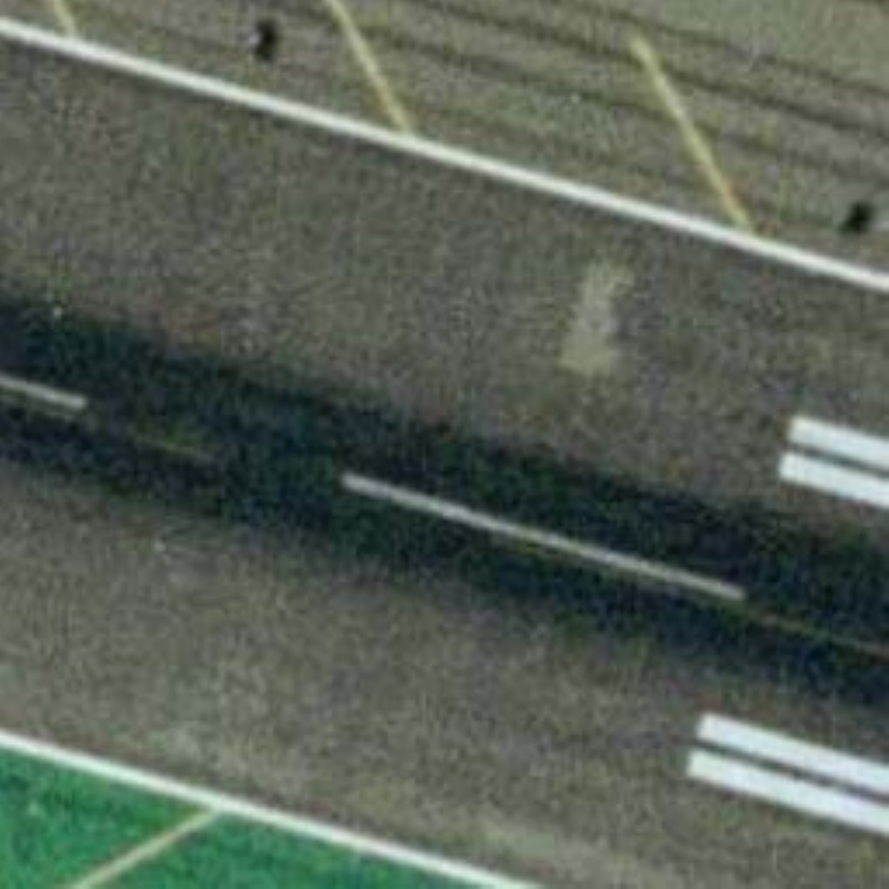}}
\subfigure[]{\includegraphics[width=0.063\textwidth]{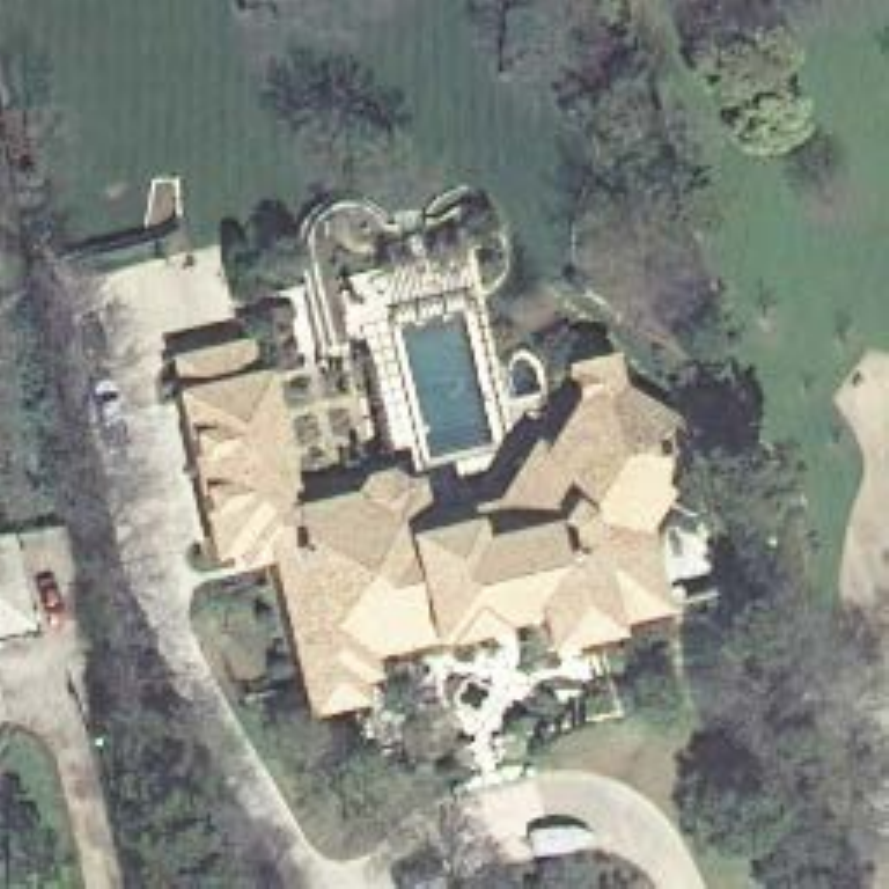}}
\subfigure[]{\includegraphics[width=0.063\textwidth]{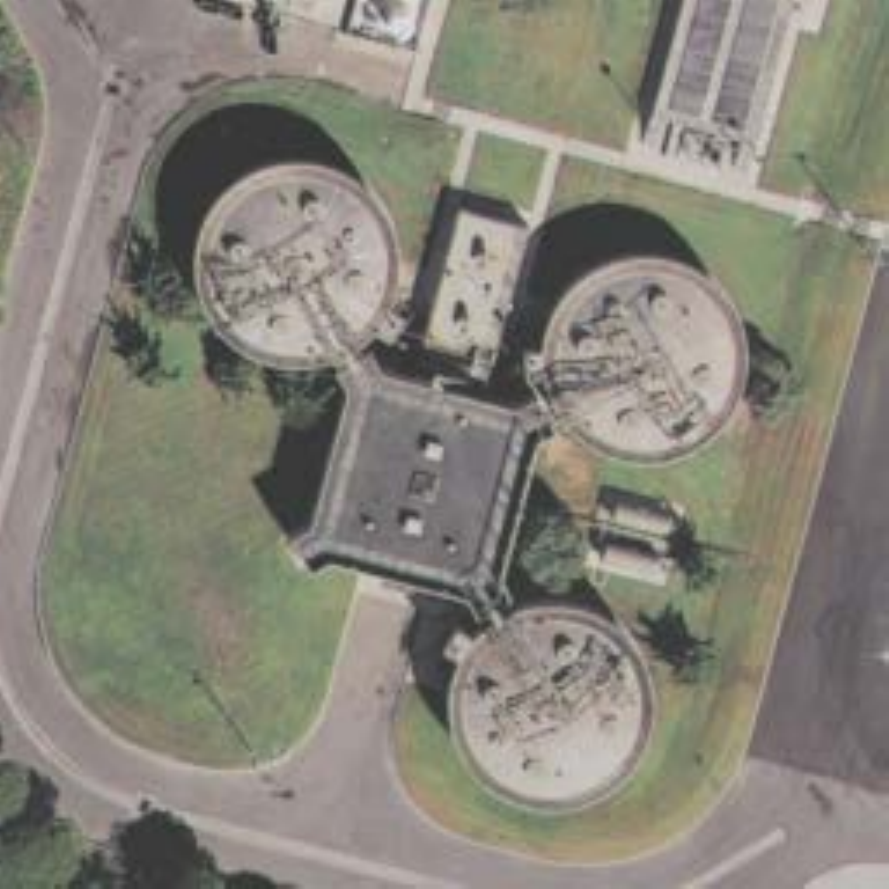}}
\subfigure[]{\includegraphics[width=0.063\textwidth]{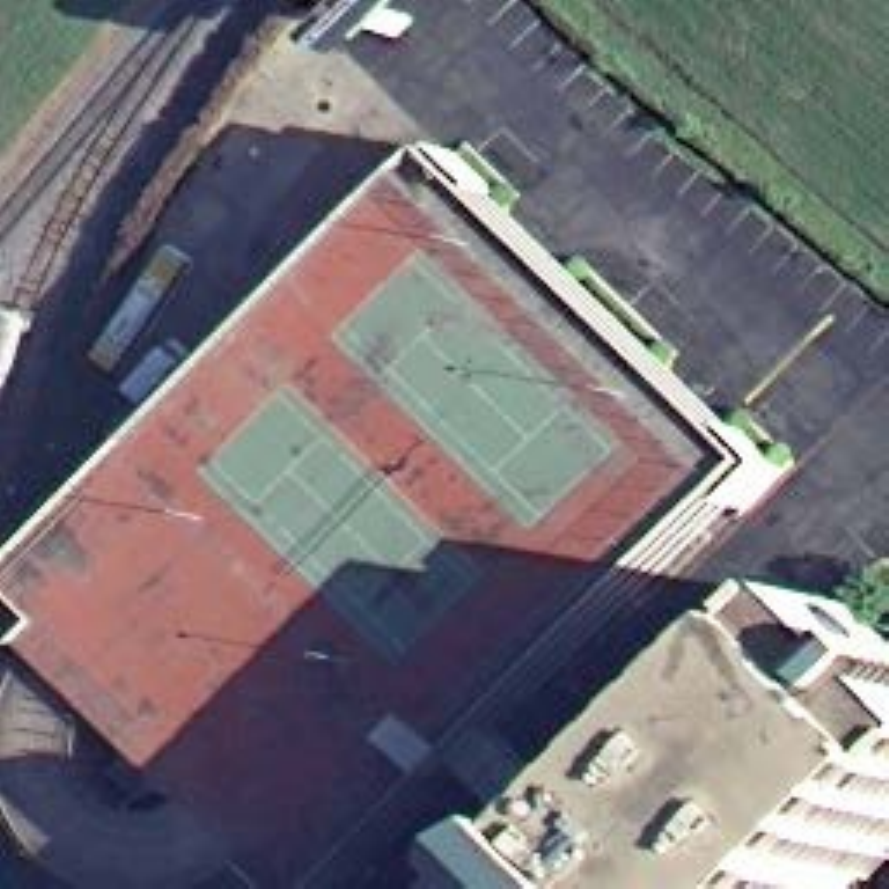}}}
\caption{Scene samples from Ucmerced Land Use dataset. (a) agricultural; (b) airplane; (c) baseball diamond; (d) beach; (e) buildings; (f) chaparral; (g) dense residential; (h) forest; (i) freeway; (j) golf course; (k) harbor; (l) intersection; (m) medium density residential; (n) mobile home park; (o) overpass; (p) parking lot; (q) river; (r) runway; (s) sparse residential; (t) storage tanks; (u) tennis court.}
\label{fig:ucmerced}
\end{figure}

\begin{figure}[htbp]
\centerline{
\subfigure[coffee]{{\includegraphics[width=0.063\textwidth]{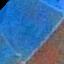}}
{\includegraphics[width=0.063\textwidth]{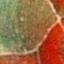}}
{\includegraphics[width=0.063\textwidth]{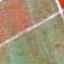}}
{\includegraphics[width=0.063\textwidth]{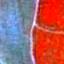}}
{\includegraphics[width=0.063\textwidth]{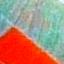}}
{\includegraphics[width=0.063\textwidth]{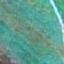}}
{\includegraphics[width=0.063\textwidth]{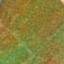}}}}
\centerline{
\subfigure[noncoffee]{{\includegraphics[width=0.063\textwidth]{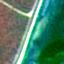}}
{\includegraphics[width=0.063\textwidth]{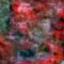}}
{\includegraphics[width=0.063\textwidth]{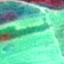}}
{\includegraphics[width=0.063\textwidth]{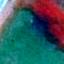}}
{\includegraphics[width=0.063\textwidth]{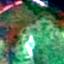}}
{\includegraphics[width=0.063\textwidth]{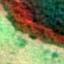}}
{\includegraphics[width=0.063\textwidth]{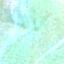}}}}
\caption{Samples of different classes from Brazilian Coffee Scene dataset.}
\label{fig:coffee}
\end{figure}

To further validate the effectiveness of the proposed method, we conduct experiments over the Ucmerced Land Use dataset \cite{ucmerced} and the Brazilian Coffee Scene dataset \cite{brazilian}. The Ucmerced Land Use dataset consists of 2100 high resolution aerial scenes (1 foot per pixel) with $256\times 256$ pixels which can be divided into 21 classes.
The Brailian Coffee Scene dataset contains 2876 multi-spectral scenes with $64\times 64$ pixels which can be divided into 2 classes.
Fig. \ref{fig:ucmerced} and \ref{fig:coffee} show the samples from the Ucmerced Land Use dataset and the Brazilian Coffee Scene dataset, respectively.

The deep model is implemented on Caffe which is a commonly used deep learning framework (see \cite{caffe} for details). CaffeNet is chosen as the deep CNN model to extract unsupervised features from the remote sensing scenes. It should be noted that in the experiments, the dimension of the last fully-connected layer is set to 512 to decrease the parameters in the model and accelerate the training process. In addition, the learning rate, the training epoch are set to 0.00001, 10000, respectively.

With the proposed method, feature vectors of the scene images are obtained. To evaluate the performance of the obtained features, we choose SVM classifier to predict scene labels with the obtained features, and LSSVM is adopted \cite{svm}.
In the experiments, both the datasets have been equally divided into five folds. To accurately validate the performance of the proposed method, all the results are obtained from the average and the standard deviation of the five-fold cross-validation.

\subsection{Results Over the Ucmerced Land Use dataset}

\begin{figure}[htbp]
\centerline{
{{\includegraphics[width=0.48\textwidth]{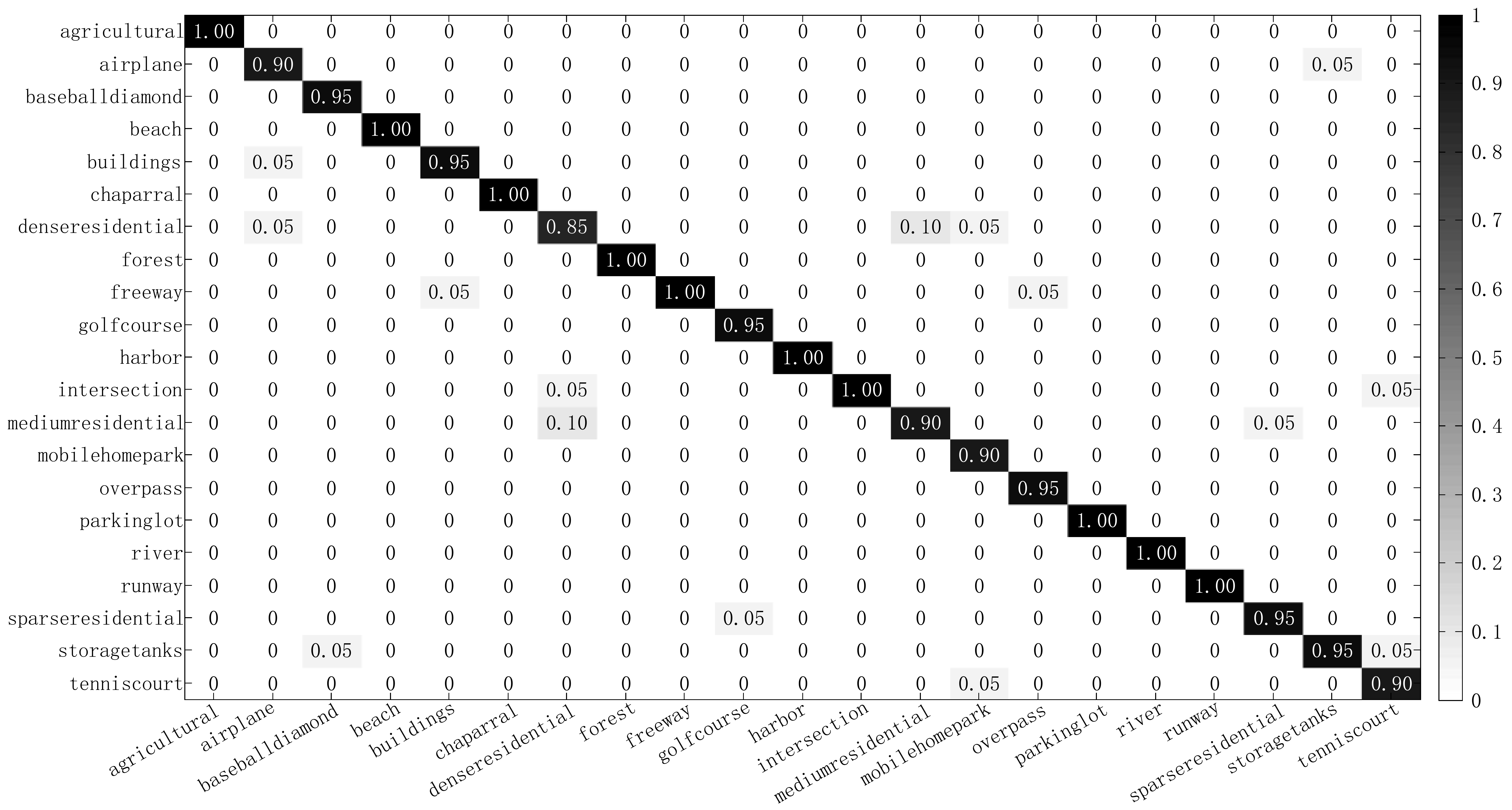}}}}
\caption{Confusion matrix of the proposed method over the Ucmerced Land Use dataset.}
\label{fig:ucmerced_confusion}
\end{figure}

Through experiments over the Ucmerced Land Use dataset,  the classification accuracy can achieve $94.33\%$ with the proposed method. The confusion matrix can be seen in Fig. \ref{fig:ucmerced_confusion}. From the confusion matrix, we can find that only some classes with great similarity could not be separated with the proposed method, such as the denseresidential and the mediumresidential, the mediumresidential and the sparseresidential. The classification errors of denseresidential/mediumresidential and mediumresidential/denseresidential can be 10\% and the error of sparseresidential/mediumresidential is 5\%. Most of the classes can be discriminated. The results show the effectiveness of the proposed method for unsupervised learning of remote sensing scenes. However, the classification performance of the proposed method can be affected by the hyper-parameter $\lambda$ and the number of pseudo classes.

\subsubsection{Classification Performance with Different Hyper-parameter $\lambda$}

As subsection \ref{subsec:proposed} shows, the $\lambda$ denotes the tradeoff between the pseudo softmax loss and the pseudo center loss. The pseudo center loss have significant effects on the update of the center point of each pseudo class and therefore the classification performance can be significantly affected by the hyper-parameter $\lambda$.

\begin{figure}[htbp]
\centerline{
{{\includegraphics[width=0.45\textwidth]{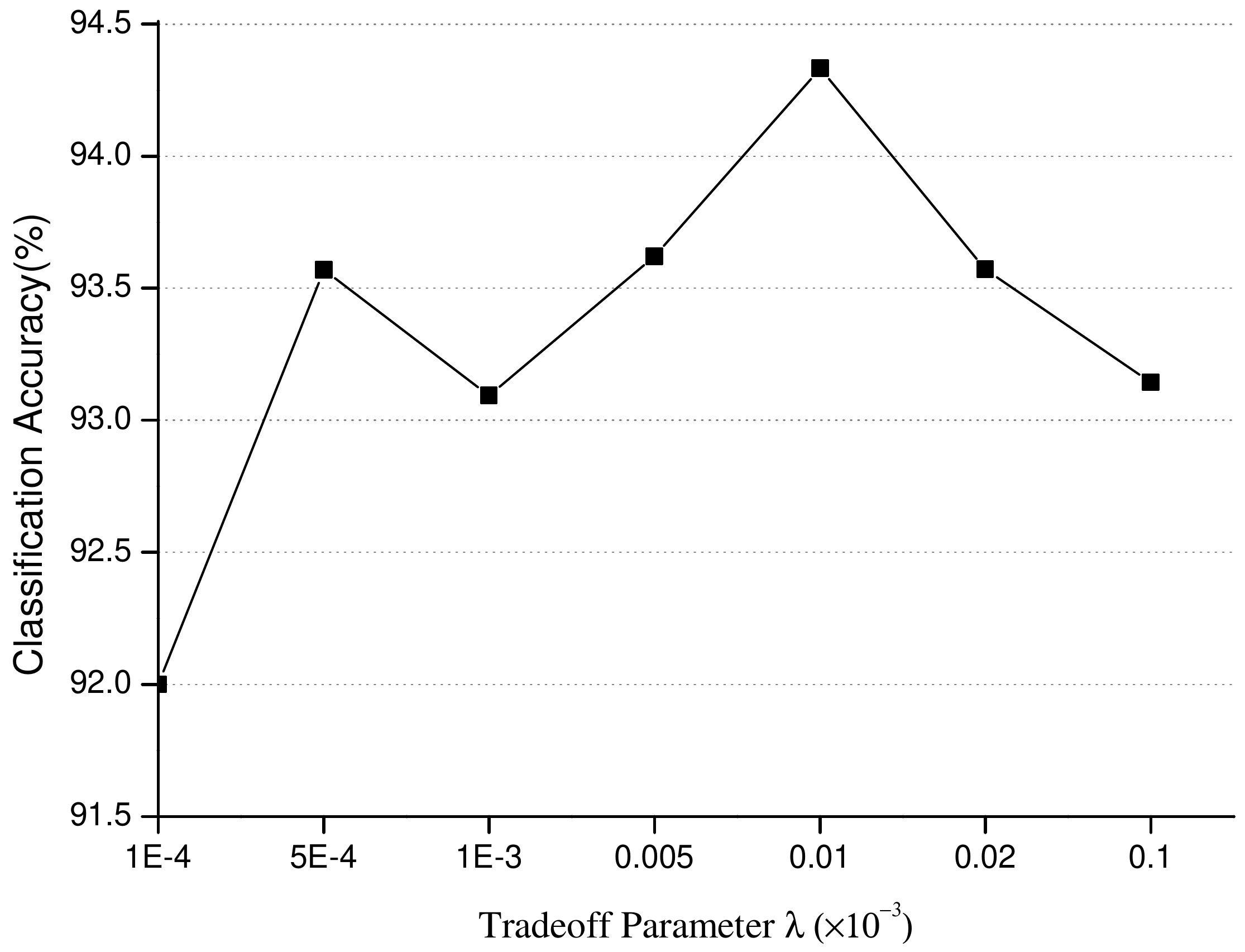}}}}
\caption{Classification accuracy obtained by the proposed method with different tradeoff parameter $\lambda$ over Ucmerced Land Use dataset.}
\label{fig:ucmerced_lambda}
\end{figure}

Fig. \ref{fig:ucmerced_lambda} presents the classification performance of the proposed method with different $\lambda$ over Ucmerced Land Use. The results are obtained when the pseudo classes is set to 10. We can find from the tendency of accuracies with different $\lambda$ in Fig. \ref{fig:ucmerced_lambda} that with the increase of the value of $\lambda$, the center points of the pseudo classes can be fully learned and be more accurate to describe the unlabelled data. Therefore, the classification performance is improved. However, when the lambda is extensively large, the training process focuses too much attention on the update of the center points which may cause the decrease of the classification performance. In addition, from Fig. \ref{fig:ucmerced_lambda}, we can find that the performance can obtain $94.33\% \pm 1.06\%$ which ranks the best when $\lambda$ is set to $10^{-5}$.

\subsubsection{Classification Performance with Different Number of Pseudo-Classes}

In the experiments, we choose 2, 5, 10, 15, 21 as the number of pseudo-classes for the proposed method over the Ucmerced Land Use dataset, respectively. The number of pseudo classes has obviously effects on the classification performance of the proposed method.

\begin{figure}[htbp]
\centerline{
{{\includegraphics[width=0.45\textwidth]{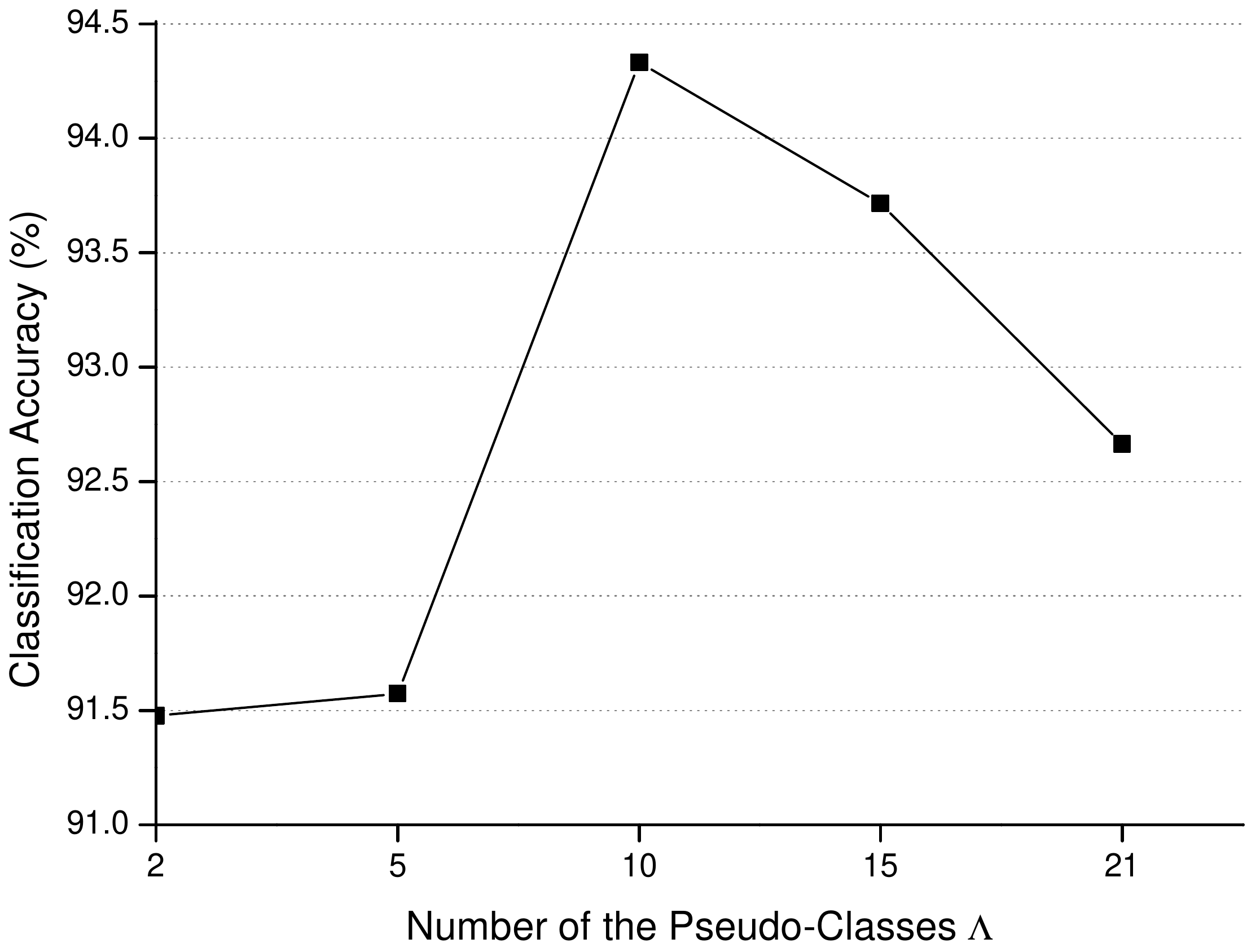}}}}
\caption{Effects of the number of pseudo-classes on the performance of the proposed method over Ucmerced Land Use dataset.}
\label{fig:ucmerced_number}
\end{figure}

Fig. \ref{fig:ucmerced_number} shows the classification accuracies of the proposed method with different number of pseudo classes over Ucmerced Land Use dataset. In the experiments, the hyper-parameter $\lambda$ is set to $10^{-4}$. We can find that with the increase of the pseudo classes, the classification performance is improved and too many pseudo classes would decrease the performance. Too small pseudo classes would make samples from different classes be assigned to the same class, and therefore the learned model could not separate different samples. In contrast, too many pseudo classes would make samples from the same class be assigned to different pseudo classes, which would also decrease the classification performance. From Fig. \ref{fig:ucmerced_number}, it can be noted that the classification performance can achieve $94.33\%$ when the number of pseudo-classes is set to 10.

\subsubsection{Comparisons with the Most Recent Methods}

To comprehensively validate the effectiveness of the proposed method for unsupervised learning of remote sensing scene representation, we compare the proposed method with other state-of-the-art methods. Table \ref{table:comparison_uc} lists the classification accuracies of several state-of-the-art unsupervised learning methods over Ucmerced Land Use dataset. From the table, we can find that the proposed method which can obtain $94.33\%$ outperforms other hand-crafted features, such as Dense SIFT (81.67\%) \cite{b5}, SPCK++ (76.05\%) \cite{b6}, UFL-SC (90.26\%) \cite{b4}, and COPD (91.33\%) \cite{b7}. In addition, when compared with other deep models, such as MARTA GANs without data augmentation (85.37\%) \cite{b1, b3}, CNN-1 (84.53\%)\cite{b2} and UCFFN (88.57\%) \cite{b1}, the proposed method can also obtain better performance. Therefore, the proposed method can obtain comparable or even better performance over the Ucmerced Land Use dataset when compared with other state-of-the-art methods, including the hand-crafted feature-based methods, and deep methods.

\begin{table}[htbp]
\caption{Comparisons with  other recent methods for unsupervised learning over Ucmerced Land Use dataset.}
\begin{center}
\begin{tabular}{c c}
\hline
\textbf{Methods} & \textbf{Accuracy(\%)} \\
\hline
Dense SIFT \cite{b5}   &  ${{81.67 \pm 1.23}}$ \\
SPCK++  \cite{b6}   &  ${{76.05}}$ \\
UFL-SC  \cite{b4}   &  ${{90.26 \pm 1.51}}$ \\
COPD  \cite{b7}   &  ${{91.33 \pm 1.11}}$ \\
{MARTA GANs} (without data augmentation) \cite{b1, b3} &  ${{85.37}}$ \\
{CNN-1}  \cite{b2} &  ${{84.53}}$ \\
{UCFFN} \cite{b1} &  $88.57$ \\
\textbf{Proposed Method} &  $\mathbf{{94.33 \pm 1.06}}$ \\
\hline
\end{tabular}
\label{table:comparison_uc}
\end{center}
\end{table}

\subsection{Results Over the Brazilian Coffee Scene dataset}

\begin{figure}[htbp]
\centerline{
{{\includegraphics[width=0.45\textwidth]{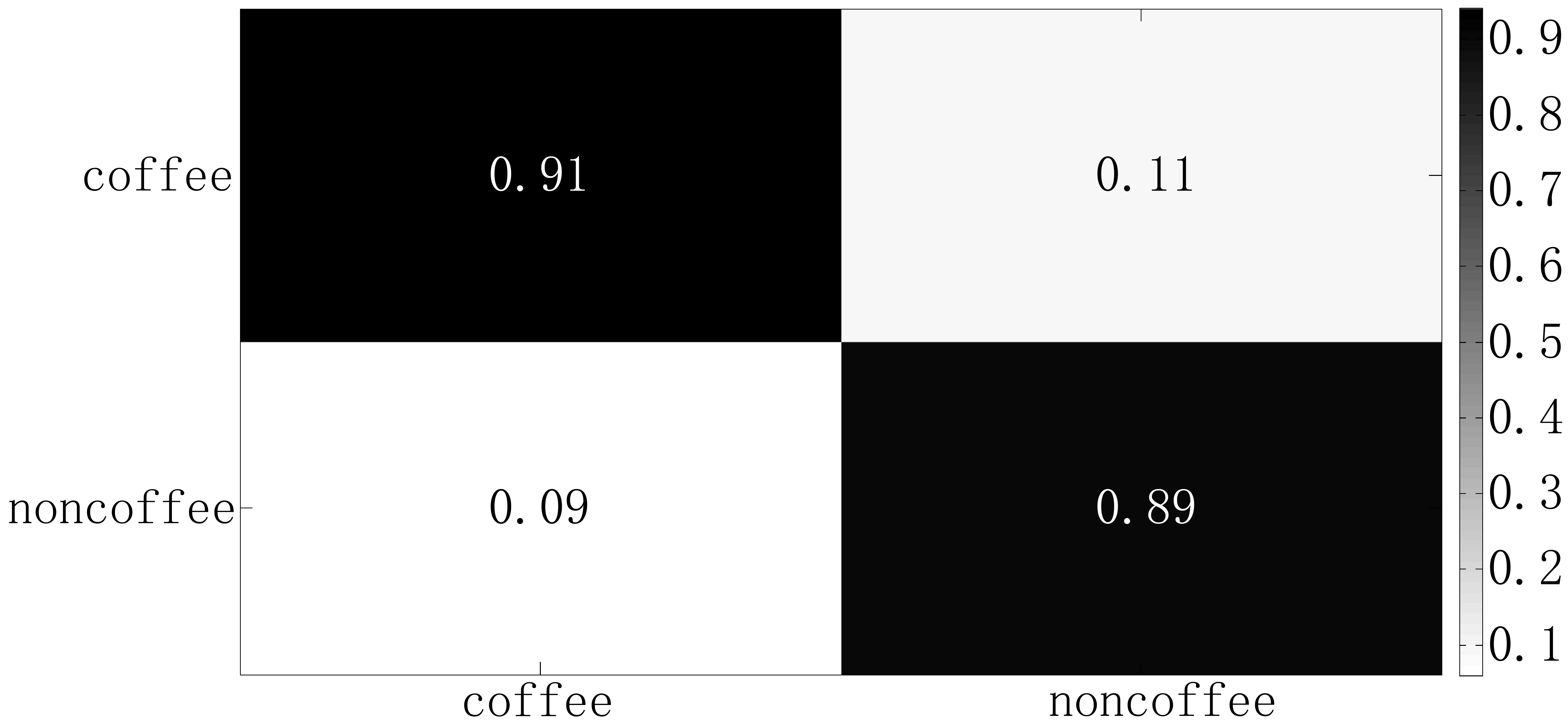}}}}
\caption{Confusion matrix of the proposed method over the Brazilian Coffee Scene dataset.}
\label{fig:brazilian}
\end{figure}

The proposed method can obtain $87.74\%+1.59\%$ over Brazilian Coffee Scene dataset. The corresponding confusion matrix is shown in Fig. \ref{fig:brazilian}. From the confusion matrix, we can find that the classification errors of coffee/noncoffee, and noncoffee/coffee are 9\% and 11\%, respectively. Obviously, the classification performance of the proposed method can be significantly affected by $\lambda$ and the number of pseudo classes.

\subsubsection{Classification Performance with Different Hyper-parameter $\lambda$}

\begin{figure}[htbp]
\centerline{
{{\includegraphics[width=0.45\textwidth]{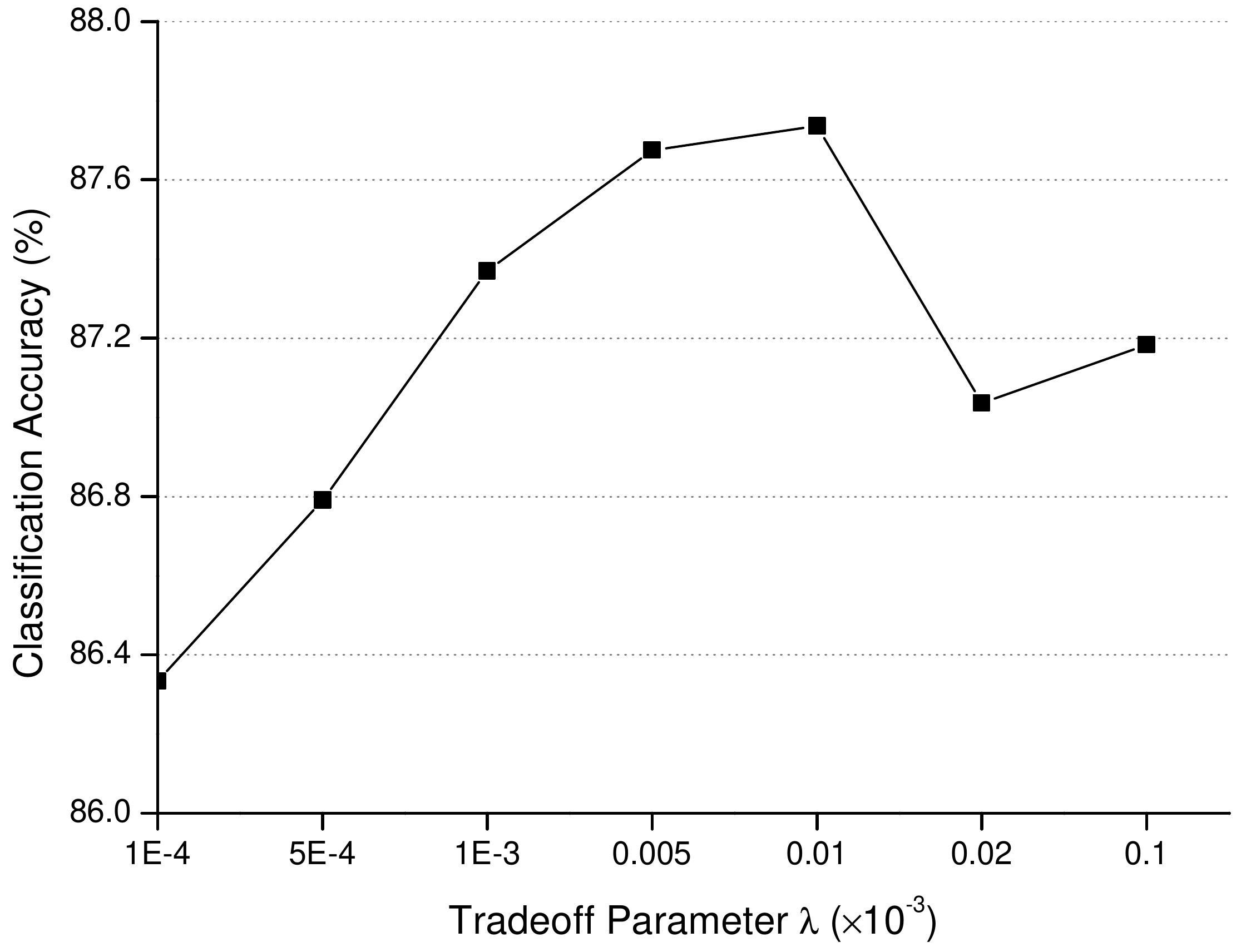}}}}
\caption{Classification accuracy obtained by the proposed method with different tradeoff parameter $\lambda$ over Brazilian Coffee Scene dataset.}
\label{fig:brazilian_lambda}
\end{figure}

Over Brazilian Scene dataset, we also investigate the results with the sets of $\lambda$ as $\{10^{-7}, 5\times 10^{-7}, 10^{-6}, 5\times 10^{-6}, 10^{-5}, 2\times 10^{-5}, 1\times 10^{-4}\}$. The classification results with different $\lambda$ are shown in Fig. \ref{fig:brazilian_lambda}. From the tendency, we can find that it is important to choose a proper $\lambda$ for the proposed method and the proposed method achieves $87.74\%$ which ranks the best when $\lambda$ is set to $10^{-5}$. It should also be noted that when $\lambda$ is extensively large, the training process may not be converged.

\subsubsection{Classification Performance with Different Number of Pseudo-Classes}
Just as the Ucmerced Land Use dataset, the number of the pseudo classes can affect the classification performance. Since the Brazilian Coffee Scene dataset contains two classes, this work conducts experiments over the Brazilian Coffee Scene dataset where the number of pseudo-classes is chosen from \{2,3,4,5,6,7,8\}.

\begin{figure}[htbp]
\centerline{
{{\includegraphics[width=0.45\textwidth]{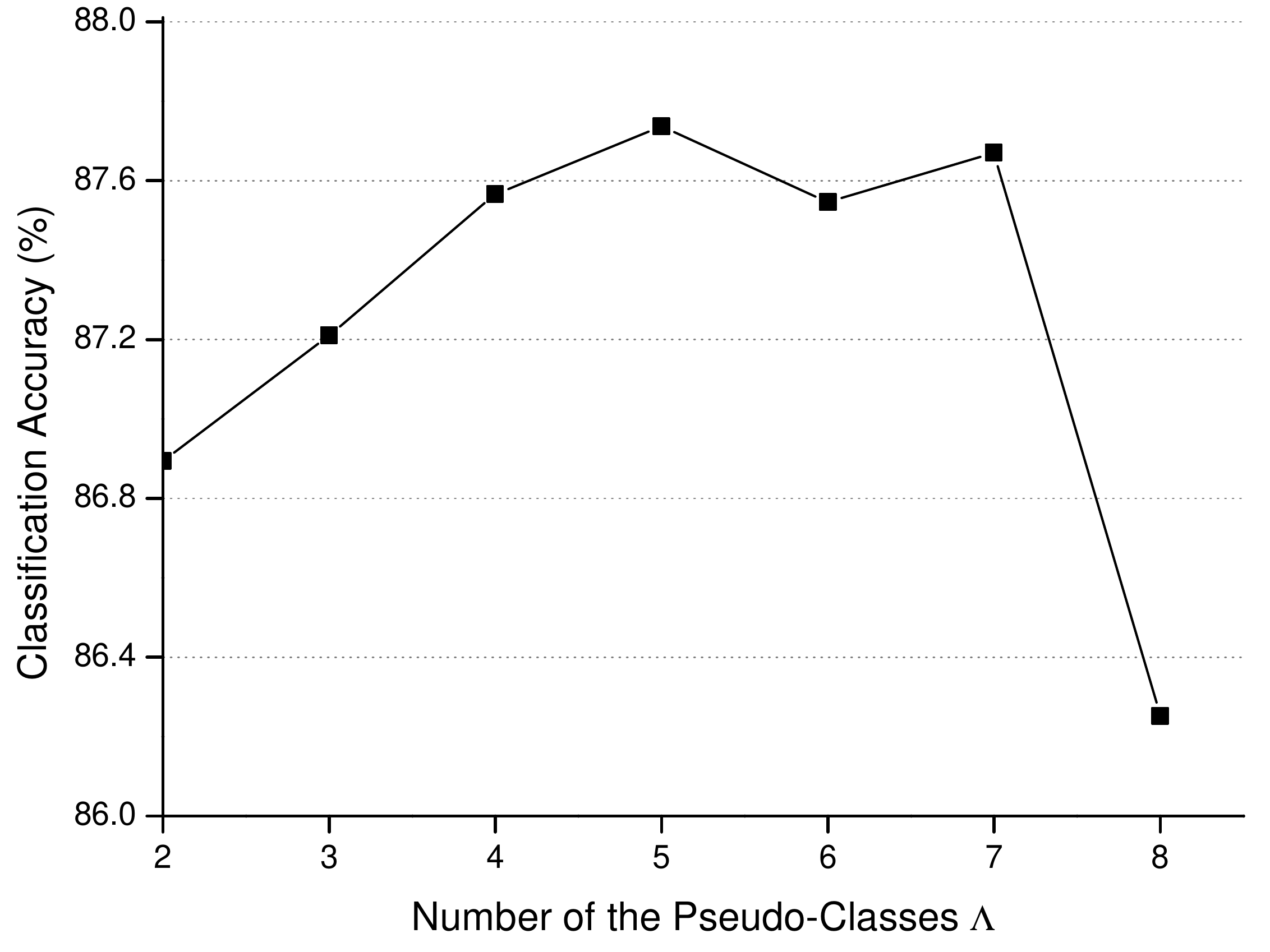}}}}
\caption{Effects of the number of pseudo-classes on the performance of the proposed method over Brazilian Coffee Scene dataset.}
\label{fig:brazilian_number}
\end{figure}

Fig. \ref{fig:brazilian_number} shows the classification performance of the proposed method with different number of pseudo-classes with the hyper-parameter $\lambda=1\times 10^{-5}$. We can find from the figure that the proposed method achieves $87.74\%$ which ranks the best when the number of pseudo classes is set to 5.

\subsubsection{Comparisons with the Most Recent Methods}

Table \ref{table:comparison_brazilian} lists the classification accuracies of several state-of-the-art methods over the Brazilian Coffee Scene dataset. The proposed method which can obtain $87.74\%$ outperforms other shallow methods, such as SIFT (82.83\%) \cite{gong}, BIC (87.03\%) \cite{brazilian}, BOVW (80.50\%) \cite{brazilian}, and OverFeat$_L$+OverFeat$_S$ (83.04\%) \cite{brazilian}. When compared with other deep models, the proposed method can obtain comparable results. The proposed method can obtain $87.74\%$ which is better than $86\%$ obtained by CNN-1 \cite{b2} and $87.69\%$ by MARTA GANs without data augmentation \cite{b1, b3}. It can obtain comparable results when compared with UCFFN (87.83\%). The comparisons show the superiority of the proposed method on unsupervised learning of the remote sensing scenes.

\begin{table}[htbp]
\caption{Comparisons with other recent methods for unsupervised learning over Brazilian Coffee dataset.}
\begin{center}
\begin{tabular}{c c}
\hline
\textbf{Methods} & \textbf{Accuracy(\%)} \\
\hline
SIFT \cite{gong} &  $82.83$ \\
BIC \cite{brazilian} &  $87.03 \pm 1.07$ \\
BOVW \cite{brazilian} &  $80.50$ \\
OverFeat$_L$+OverFeat$_S$ \cite{brazilian} &  $83.04 \pm 2.00$ \\
CNN-1 \cite{b2} &  $86.00$ \\
MARTA GANs (without data augmentation) \cite{b1, b3} & $87.69$ \\
UCFFN \cite{b1} &  $87.83$ \\
\textbf{Proposed Method} &  $\mathbf{{87.74 \pm 1.59}}$ \\
\hline
\end{tabular}
\label{table:comparison_brazilian}
\end{center}
\end{table}

\section{Conclusions}

This work proposes a novel end-to-end unsupervised learning method for the representation of remote sensing scenes. First, the proposed method chooses the CNN model to extract features from the scenes. Then, center points are introduced in the training process to formulate the pseudo classes. By allocating pseudo labels to different samples based on the center points and formulating the pseudo center loss to decrease the variance between the samples and the corresponding center point, different samples can be clustered with the center points. In addition, through joint learning of the pseudo center loss and the pseudo softmax loss, the center points and the parameters in CNN model are both updated. Experimental results show that the proposed end-to-end unsupervised learning process can extract discriminative features from the scenes. In addition, the proposed method can obtain comparable or even better results when compared with other state-of-the-art methods.

In future work, we intend to apply the proposed unsupervised learning methods on other computer vision tasks, such as visual representation.  In addition,  we also would like to evaluate the performance of the proposed method with other CNN model, such as GoogLeNet, ResNet. Moreover, other technologies, which can improve the performance of the unsupervised learning methods, would be another interesting topic.

\vspace{12pt}

\end{document}